\documentclass[journal]{IEEEtran}

\usepackage{multirow,booktabs,color,soul,threeparttable}
\definecolor{hl}{rgb}{0.75,0.75,0.75}
\sethlcolor{hl}
\usepackage{graphicx} 
\usepackage{amsmath}
\usepackage{algorithm}
\usepackage{algorithmic} 
\usepackage{marvosym}
\usepackage{makecell}

\begin{document}
\title{A Detection Region Method-Based Evolutionary Algorithm for Binary Constrained Multiobjective Optimization
}

\author{
	Weixiong Huang,
	Rui Wang*, \emph{Senior Member}, 
	Tao Zhang,
	Sheng Qi,
	and Ling Wang,  \emph{Senior Member}
	\thanks{This work has been submitted to the IEEE for possible publication. Copyright may be transferred without notice, after which this version may no longer be accessible.}
	\thanks{Weixiong Huang, Rui Wang, Tao Zhang, and Sheng Qi are with the College of Systems Engineering, National University of Defense Technology, Changsha, 410073, China, and the Hunan Key Laboratory of Multi-energy System Intelligent Interconnection Technology, Changsha, 410073, China. Rui Wang is also with Xiangjiang Laboratory, Changsha, 410205, China. (e-mail: huangweixiong@nudt.edu.cn; ruiwangnudt@gmail.com; zhangtao@nudt.edu. cn; qisheng@nudt.edu. cn).}
	\thanks{Ling Wang is with the Department of Automation, Tsinghua University, Beijing, 100084, China (e-mail: wangling@tsinghua.edu.cn).}
	\thanks{Corresponding Author: Rui Wang (Email:ruiwangnudt@gmail.com).}
}
\maketitle

\begin{abstract}
Solving constrained multi-objective optimization problems is highly challenging, especially when constraint functions are unknown or unquantifiable, yielding only binary outcomes (feasible or infeasible). Such cases significantly limit the performance of many advanced constrained multi-objective evolutionary algorithms (CMOEAs), particularly those relying on \(\varepsilon\)-based methods. To address these challenges, this paper proposes a novel CMOEA framework based on detection region method (DRMCMO). In DRMCMO, detection regions dynamically monitor feasible solutions to enhance convergence. By using existing feasible solutions as a reference, the detection regions are dynamically adjusted to help the population escape local optima. We have modified three existing test suites to serve as benchmark problems for Constrained Multi-Objective Problems with Binary Constraints (CMOP/BC) and conducted comprehensive comparative experiments with state-of-the-art algorithms on these test suites and real-world problems. The results demonstrate that DRMCMO is highly competitive against other state-of-the-art algorithms. Given the limited research on CMOP/BC, our study provides a new perspective for advancing this field.
\end{abstract}


%
\IEEEpeerreviewmaketitle

\section{Introduction}
%
%
%
%
\IEEEPARstart{M}{any} engineering applications involve multiple conflicting optimization objectives and one or more constraints. For example, testing resource allocation problems\cite{su2021enhanced}, web service location-allocation problems\cite{tan2018evolutionary}, and hybrid renewable energy system optimization \cite{chen2022constraint} all fall into this category. Solving these problems requires balancing multiple objectives while ensuring all constraints are satisfied, making the problem-solving process more complex and challenging. These problems can be classified as constrained multiobjective optimization problems (CMOPs).

In practical optimization scenarios, a significant category of constraints often remains unknown or unquantifiable, thereby adding substantial complexity to the problem-solving process. For instance, in autonomous driving, collision avoidance is a crucial constraint that typically provides only binary feedback: either a collision is avoided (feasible) or it occurs (infeasible) \cite{garcia2015comprehensive}. Similarly, in the Job Shop Scheduling Problem, strict precedence relationships between operations must be maintained \cite{brailsford1999constraint}; failure to satisfy these constraints renders the solution infeasible. In multi-robot path planning, constraints that prevent collisions between robots also offer only binary feedback, with no quantifiable measures of constraint violation \cite{ryan2010constraint}.
Non-quantifiable constraints are prevalent in CMOPs and demand meticulous consideration. Recently, Li et al. \cite{li2024evolutionary} identified a subclass of CMOPs with unknown constraints, termed CMOP/UC. This subclass essentially involves binary constraints, where the constraints can only return two possible outcomes: 0 (satisfied) or 1 (violated). For clarity and consistency, this paper refers to this problem as CMOP/BC, short for "Constrained Multiobjective optimization Problems with Binary Constraints."

Over the past two decades, evolutionary algorithms(EAs) have significantly advanced in solving multiobjective optimization problems \cite{coello2002theoretical}. These multiobjective optimization evolutionary algorithms(MOEAs) can be broadly categorized into three types: 1) Dominance-based algorithms:  Examples include NSGA-II \cite{deb2002fast} and SPEA2 \cite{zitzler2001spea2}, which excel at problems with conflicting objectives and produce well-distributed Pareto-optimal solutions. 2) Decomposition-based algorithms: Algorithms such as MOEA/D \cite{zhang2007moea} and MSOPS-II \cite{hughes2007msops} are advantageous in many objectives problems, providing satisfactory Pareto sets with limited resources. 3) Indicator-based algorithms: Algorithms such as IBEA \cite{zitzler2004indicator} and HypE \cite{bader2011hype} use performance indicators to guide the selection process towards well-distributed and high-quality solutions. Each algorithm type has its strengths and weaknesses, making it essential for researchers to select or hybridize them based on problem characteristics.

However, these MOEAs cannot directly solve CMOPs as they lack essential components for handling constraints \cite{liang2022survey}. Therefore, various effective constraint-handling techniques (CHTs) have been developed over the past few decades to enable MOEAs to tackle CMOPs. These CHTs can be broadly categorized into the following approaches: 1) Feasibility-oriented methods \cite{deb2002fast} \cite{fonseca1998multiobjective}: These methods prioritize feasible solutions during the selection process, which may result in excessive selection pressure on feasible solutions. 2) \(\varepsilon\)-based methods \cite{takahama2010efficient} \cite{takahama2010constrained}: These methods solve CMOPs by adjusting constraint boundaries, though selecting an appropriate \(\varepsilon\) variant can be challenging. 3) Multi-objective optimization-based methods \cite{wang2012combining} \cite{jiao2021two}: Some methods reformulate the CMOP as a multiobjective optimization problem, treating constraint violations as additional objectives, which can complicate objective optimization. 4) Stochastic ranking \cite{runarsson2000stochastic}: This approach probabilistically favors individuals with superior objective convergence while occasionally selecting those with better constraint violations. However, not all solutions exhibit dominance relationships, and determining these probabilities requires careful consideration. 5) Penalty functions \cite{jan2013study} \cite{joines1994use}: These algorithms incorporate constraint information into the fitness function, penalizing infeasible solutions, but tuning the penalty parameters can be challenging. These fundamental CHTs can be integrated into different MOEA frameworks. For example, multi-population or multi-stage constrained MOEAs may employ a combination of these techniques to effectively solve complex CMOPs. By leveraging these advanced CHTs, MOEAs have become increasingly capable of addressing various real-world optimization problems with diverse constraints.

In CMOP/BC, however, the binary nature of constraint outcomes complicates the accurate quantification of constraint violations (CV) \cite{li2024evolutionary}. This limitation may lead to a decline in the performance of many otherwise effective CHTs and CMOEAs. For instance, in CMOP/BC problems, \(\varepsilon\)-based methods struggle to scale the feasible region, resulting in significant performance drops. It is also worth noting that research on CMOP/BC is currently limited, with few practical algorithms available. To promote further research in this area, this paper proposes a constrained multiobjective evolutionary algorithm framework based on the detection region method (DRMCMO). The main contributions of DRMCMO are as follows:

  \begin{enumerate}[]
  	\item Proposing the Detection Region Method (DRM): We introduce a detection region method to address the challenges of solving CMOP/BC, particularly the limitations of \(\varepsilon\)-based methods. This method dynamically relaxes the selection pressure on feasible solutions and monitors promising feasible solutions to assist the population in escaping local optima.
  	
  	\item Adapting Existing CMOP Test Problems:  In light of the limited research on CMOP/BC, we have adapted three existing CMOP benchmark test suites for conducting comparative experiments.
  	
  	\item   Outstanding Experimental Performance: Our experimental results highlight the competitiveness of the proposed algorithm, showcasing its effectiveness in addressing CMOP/BC challenges.

  \end{enumerate}

The remainder of this paper is organized as follows. Section \ref{s2} presents the foundational concepts relevant to this study and reviews current constrained multiobjective evolutionary algorithms (CMOEAs). Section \ref{s3} details the proposed algorithm and its components. Sections \ref{s4} and \ref{s5} outline the experimental settings and analyze the results of a series of experiments, respectively. Finally, Section \ref{s6} concludes the paper and discusses potential directions for future work.

\section{Related Work}
\label{s2}
\subsection{Basic Definitions}
\subsubsection{CMOP}
\begin{equation}
	\label{e1}
	\begin{cases}
		\text{minimize} \ F(\mathbf{x})=(f_{1}(\mathbf{x}),f_{2}(\mathbf{x}),\cdots,f_{m}(\mathbf{x}))^{T}
		\\ \text{subject  to} \ \  \mathbf{x} \in \Omega,
		\\ \quad  \quad \quad \quad \ \ g_{i}(\mathbf{x})\leq0, \  i=1,\cdots,p
		\\ \quad  \quad \quad \quad \ \ h_{j}(\mathbf{x})=0, \ j=1,\cdots,q
	\end{cases}
\end{equation}
where $\mathbf{x} = (x_1, x_2, \ldots, x_D)$ represents an $n$-dimensional vector of decision variables within the decision space $\Omega$. The function $\mathbf{F}(\mathbf{x}): \Omega \rightarrow \mathbf{R}^m$ denotes the objective function vector, comprising $m$ conflicting objectives. The objective space is denoted by $\mathbf{R}^m$. The constraint $g_i(\mathbf{x}) \leq 0$ represents the $i$-th inequality constraint, with $p$ being the total number of such constraints. Similarly, $h_j(\mathbf{x}) = 0$ denotes the $j$-th equality constraint, with $q$ being the total number of equality constraints.

\subsubsection{CMOP/BC}
\begin{equation}
		\label{e2}
	\begin{cases}
		\text{minimize} \ F(\mathbf{x})=(f_{1}(\mathbf{x}),f_{2}(\mathbf{x}),\cdots,f_{m}(\mathbf{x}))^{T}
		\\ \text{subject  to} \  \ \mathbf{x} \in \Omega,
		\\ \quad  \quad \quad \quad \ \ \mathrm{G}(\mathbf{x}) = (\underbrace{0,\ldots,0}_\ell)^\top 
		
	\end{cases}
\end{equation}
the definition of CMOP/BC is similar to that of CMOP, with the critical difference being like the constraint set $\mathbf{G}(\mathbf{x}) = (g_1(\mathbf{x}), \ldots, g_\ell(\mathbf{x}))^\top$. In CMOP/BC, each constraint can only return a binary result: $g_i(\mathbf{x})$ returns 0 if the solution $\mathbf{x}$ satisfies the constraint, and 1 if the solution $\mathbf{x}$ violates the constraint.

\subsubsection{Constraint Violation}
In CMOP, solutions assessment extends beyond the objectives to encompass the constraints. $\textit{C}_\textit{j}(\mathbf{x})$ represents the degree of violation of solution $\mathbf{x}$ on the $j$th constraint, defined as follows:
\begin{equation}
		\label{e3}
	\textit{C}_\textit{j}(\mathbf{x})=\left\{\begin{array}{l}
		\max \left(0, g_j(\mathbf{x})\right)=0, j=1, \ldots, p \\
		\max \left(0,\left|h_j(\mathbf{x})\right|-\delta\right)=0, j=p+1, \ldots, p+q
	\end{array}\right.
\end{equation}	
where $\delta$ signifies a minimal threshold (commonly set at $\delta=10^{-6}$), which is employed to marginally alleviate the strictness of equality constraints, thereby transforming them into inequality constraints\cite{cai2006multiobjective}.
The violation of $\mathbf{x}$  on all constraints can be calculated as:
\begin{equation}
	\label{e4}
	\textit{CV}(\mathbf{x})=\sum_{j=1}^{p+q} \textit{C}_\textit{j}(\mathbf{x})
\end{equation}
A solution $\mathbf{x}$ is deemed feasible if it yields a total constraint violation of $\textit{CV}(\mathbf{x})=0$; conversely if this condition is not met, $\mathbf{x}$ is considered infeasible.

Additionally, in CMOP/BC, since each constraint can only return 0 or 1, the calculation of constraint violation results in an integer. This integer represents the number of constraints violated by the solution $\mathbf{x}$.

\subsubsection{Feasible Region}
 A solution $\mathbf{x}$  satisfies all constraints, i.e., $\textit{CV}(\mathbf{x})=0$. On the contrary, if this criterion is not satisfied, the solution is classified as infeasible. The collection of all feasible solutions constitutes what is known as the feasible region.
\begin{equation}
		\label{e5}
	S=\{\mathbf{x}\mid \textit{CV}(\mathbf{x})=0, \mathbf{x} \in 	\Omega\}.
\end{equation}

\subsubsection{Pareto Dominance}
Given two solutions $\mathbf{x}_1$, $\mathbf{x}_1$, $\mathbf{x}_1$ $\prec$ $\mathbf{x}_2$ if and only if
\begin{equation}
		\label{e6}
	\\\begin{cases}\forall i\in(1,2,\ldots,m):f_i(\mathbf{x_1})\leq f_i(\mathbf{x_2})\\\exists j\in(1,2,\ldots,m):f_j(\mathbf{x_1})<f_j(\mathbf{x_2})\end{cases}
\end{equation}	
where M is the number of objectives and $\mathbf{x}_1$ $\prec$ $\mathbf{x}_2$ denotes that $\mathbf{x}_1$ dominates $\mathbf{x}_2$.

\subsubsection{Constraint Dominance Principle}
For two solutions $\mathbf{x}$ and $\mathbf{y}$, if $\mathbf{x}$ is said to dominate $\mathbf{y}$ based on constraint dominance principle (CDP), it needs to satisfy the  following conditions:
\begin{equation}
		\label{e7}
	\left\{\begin{array}{l}
		\textit{CV}(\mathbf{x})=0, \textit{CV}(\mathbf{y})>0 \\
		\textit{CV}(\mathbf{x})=0, \textit{CV}(\mathbf{y})=0, \text { and } \mathbf{x} \prec \mathbf{y} \\
		\textit{CV}(\mathbf{x})>0, \textit{CV}(\mathbf{y})>0, \text { and } \textit{CV}(\mathbf{x})<\textit{CV}(\mathbf{y})
	\end{array}\right.
\end{equation}

\subsection{Constrained Multiobjective Evolutionary Algorithms}
In the past two decades, many representative CMOEAs have been proposed, which can be roughly divided into the following categories:
\subsubsection{Feasibility-Guided CMOEAs}
These algorithms primarily focus on feasible solutions, as infeasible solutions ultimately hold no value for decision-makers. Therefore, it is logical to utilize feasible solutions to guide the search of CMOEAs. These techniques inherently prioritize feasible solutions (or place greater emphasis on them) over infeasible ones, thereby ensuring that feasible solutions drive the optimization process. Coello et al. \cite{coello1999moses} introduced a method that disregards infeasible solutions. While this approach is straightforward and easy to implement, it lacks sufficient selection pressure when all solutions are infeasible. Deb et al. \cite{deb2002fast} introduced the constrained dominance principle (CDP), where feasible solutions always dominate infeasible ones. Yu et al. \cite{yu2021dynamic} developed a dynamic selection method for objective functions and constraints, assisted by preferences. Ma et al. \cite{ma2019new} created a new fitness function incorporating two rankings: one based on CDP and another on Pareto dominance. Liu et al. \cite{liu2019handling} converted a CMOP into a constrained single-objective optimization problem to speed up evolution and applied the feasibility rule to solve the original CMOP. Oyama et al. \cite{oyama2007new} established a dominance relationship based on the extent of constraint violation for each solution.

\subsubsection{Constraint Relaxation-Based CMOEAs}
This class of algorithms primarily focuses on relaxing constraints to reduce the excessive selection pressure on feasible solutions. Representative methods in this category include penalty function approaches and $\varepsilon$-based methods. Essentially, these algorithms incorporate information from infeasible solutions into the evolutionary process to create a more balanced optimization process that allows for broader solution space exploration. Woldesenbet et al. \cite{woldesenbet2009constraint} introduced a method that leverages an adaptive penalty function alongside a distance measure function. Chih-Hao et al. \cite{lin2013rough} combined the penalty function with rough set theory, where the penalty coefficient adjusts adaptively based on an individual's constraint violations. Fan et al. \cite{fan2019improved} proposed dynamically adjusting the $\varepsilon$ value according to the ratio of feasible solutions to the total solutions in the current population, which is particularly effective for CMOPs with large infeasible regions. PPS \cite{FAN2019665} implements a population push-pull process by controlling the $\varepsilon$ value, essentially alternating between relaxing and tightening constraints. Zhu et al. \cite{zhu2020constrained} developed a MOEA/D variant based on a detect-and-escape strategy (MOEA/D-DAE) for CMOPs. When the population becomes trapped in local feasible regions, the constraint boundaries are expanded, allowing the population to escape from these local regions. Zhou et al.\cite{zhou2022domination} proposed a constrained shift-based density estimation strategy to transfer the relaxation of constraints to the objective modification.

\subsubsection{Multi-Population, Multi-Stage, or Multi-Task CMOEAs}

These CMOEAs are typically designed with a high degree of complexity and integration, often employing multiple CHTs. Despite their varied approaches, they all excel at leveraging auxiliary information. Liu et al. \cite{liu2019handling} developed a two-stage algorithm that initially transforms the CMOP into a constrained single objective optimization problem to expedite convergence before addressing the original multiobjective problem. In a different approach, Ma et al. \cite{ma2021multi} introduced a constraint-handling priority for MSCMO, where constraints are incrementally added and managed at various evolutionary stages. CMOEA-MS \cite{tian2021balancing} employs distinct fitness strategies throughout different evolutionary phases, giving equal priority to objectives and constraints in the first stage and prioritizing constraints over objectives in the second stage. Li et al. \cite{li2018two} proposed C-TAEA, a competitive two-archive coevolutionary CMOEA for CMOPs, which utilizes a convergence-oriented archive (CA) to prioritize feasible solutions and a diversity-oriented archive (DA) to encourage exploration in less-exploited regions. Additionally, Tian et al. \cite{tian2020coevolutionary} introduced CCMO, a dual-population-based coevolutionary framework where one population addresses the CMOP and the other tackles its unconstrained version, with significantly weaker cooperation between the two populations compared to existing coevolutionary algorithms. Ming et al. \cite{ming2021dual} presented a Dual-Population-based Evolutionary Algorithm wherein one population deals with infeasible solutions using a self-adaptive penalty function (saPF). At the same time, the other employs a feasibility-oriented approach. Zou et al. \cite{zou2023multipopulation} proposed MCCMO, a multi-population algorithm where each population manages a distinct constraint individually, integrating an inter-population cooperation mechanism to handle multiconstraint CMOPs.  Recently, CMOEAs based on evolutionary multitasking have also shown strong competitiveness. Qiao et al. \cite{qiao2022evolutionary} develop an Evolutionary Multitasking (EMT)-based framework, EMCMO, by transforming the optimization into two related tasks, utilizing knowledge transfer to enhance performance. MTCMO \cite{qiao2022dynamic} dynamically utilizes an auxiliary task and knowledge transfer to enhance algorithm performance by effectively incorporating infeasible solutions. IMTCMO \cite{qiao2023evolutionary} enhances MTCMO by introducing new search operators for local and global search, employing angle-based parent selection and multitasking technology to guide the auxiliary task population from the unconstrained Pareto front (UPF) to the constrained Pareto front (CPF). Ming et al. \cite{ming2023adaptive} introduce a general multitasking framework for CMOPs, utilizing reinforcement learning to select suitable auxiliary tasks and develop two novel algorithms based on Q-Learning and Deep Q-Learning. \\

\textbf{Remark 1: }
\textit{The aforementioned classes of CMOEAs each have their own strengths and unique characteristics when solving CMOPs. However, when confronted with CMOP/BC, they tend to exhibit varying degrees of performance degradation.} \\

\textbf{Remark 2: }
\textit{When dealing with CMOP/BC, feasibility-guided algorithms may struggle due to the diminished guiding capability of CV. This challenge is particularly pronounced in problems with small feasible regions, where the algorithm might fail to find any feasible solutions. Although this issue could be mitigated by employing specific initialization strategies, the pressure to select feasible solutions remains high once such solutions are found.}\\

\textbf{Remark 3: }
\textit{Algorithms that rely on constraint relaxation, particularly those using a dynamic threshold to relax CV, experience the most significant performance loss when addressing CMOP/BC. This is because the approach of relaxing CV becomes entirely ineffective for this class of problems. We will elaborate on this point in section \ref{s2} C}\\

\textbf{Remark 4: }
\textit{Multi-Population, Multi-Stage, and Multi-Task Methods are typically hybrid approaches that may employ different constraint-handling techniques across various stages, populations, or tasks. However, as long as they incorporate constraint relaxation or CV-guided methods at any point, they may still encounter some degree of performance decline when tackling CMOP/BC.}\\

\subsection{Current Challenges in Existing Work for CMOP/BC}
Many current CMOEAs and CHTs rely on CV information to guide their search processes. For instance, algorithms based on CDP compare the magnitude of CV to rank infeasible solutions. \(\varepsilon\)-based methods utilize a dynamic threshold to relax CV, while other algorithms may incorporate CV into their fitness functions. Additionally, some penalty function-based approaches also take constraint violations into account.

\begin{figure}[!htbp]
	\centering
	\footnotesize
	\begin{minipage}{4.3cm}
		\centering
		\includegraphics[width=4.3cm]{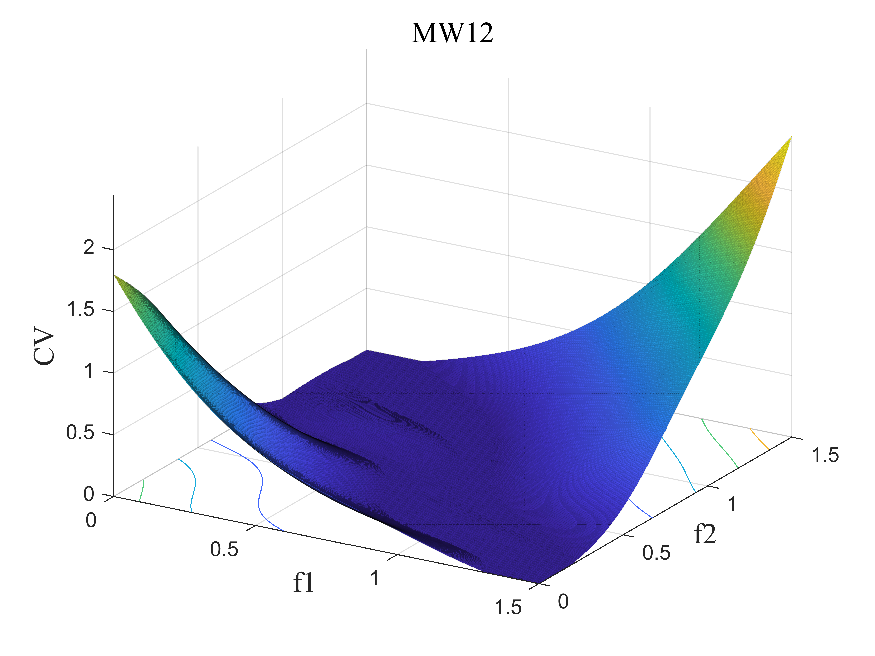}
		\text{(a)}
	\end{minipage}
	\begin{minipage}{4.3cm}
		\centering
		\includegraphics[width=4.3cm]{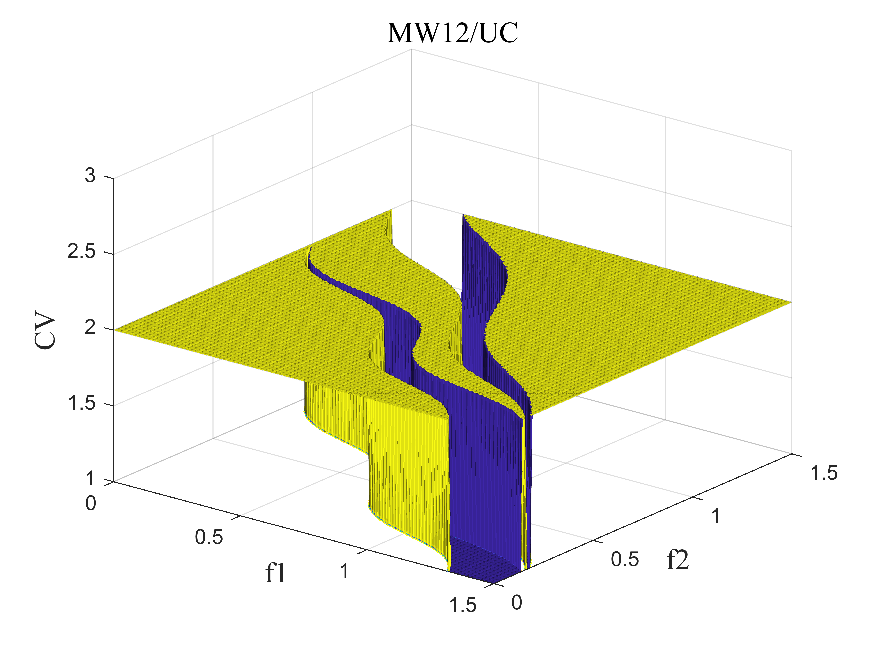}
		\text{(b)}
	\end{minipage}
	\centering
	\caption{Variation of constraint violation in the objective space for (a) MW12 and (b) MW12/BC.}
	\label{f1}
\end{figure}

However, when addressing CMOP/BC, CV may not be as effective as in traditional CMOPs. In CMOP/BC, CV merely sums the number of constraint violations, which does not accurately reflect the severity of these violations. We modified the constraints of the MW problems \cite{ma2019evolutionary} to produce binary outputs, creating the MW/BC variant. Figure \ref{f1} (a) illustrates the variation of CV in the objective space for MW12, while Fig. \ref{f1} (b) depicts the variation of CV for MW12/BC. This comparison reveals that in CMOP/BC, CV values are often identical across many regions, diminishing the guidance that CV provides for the population to locate feasible regions. Nevertheless, the selection pressure exerted by CDP on feasible solutions remains.

\begin{figure}[!htbp]
	\centering
	\footnotesize
	\begin{minipage}{4.3cm}
		\centering
		\includegraphics[width=4.3cm]{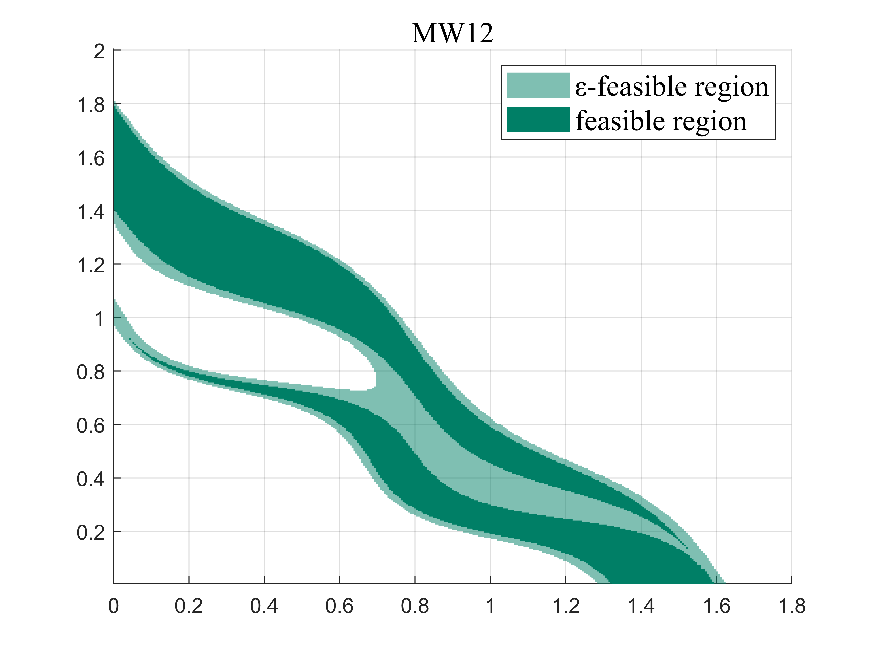}
		\text{(a)}
	\end{minipage}
	\begin{minipage}{4.3cm}
		\centering
		\includegraphics[width=4.3cm]{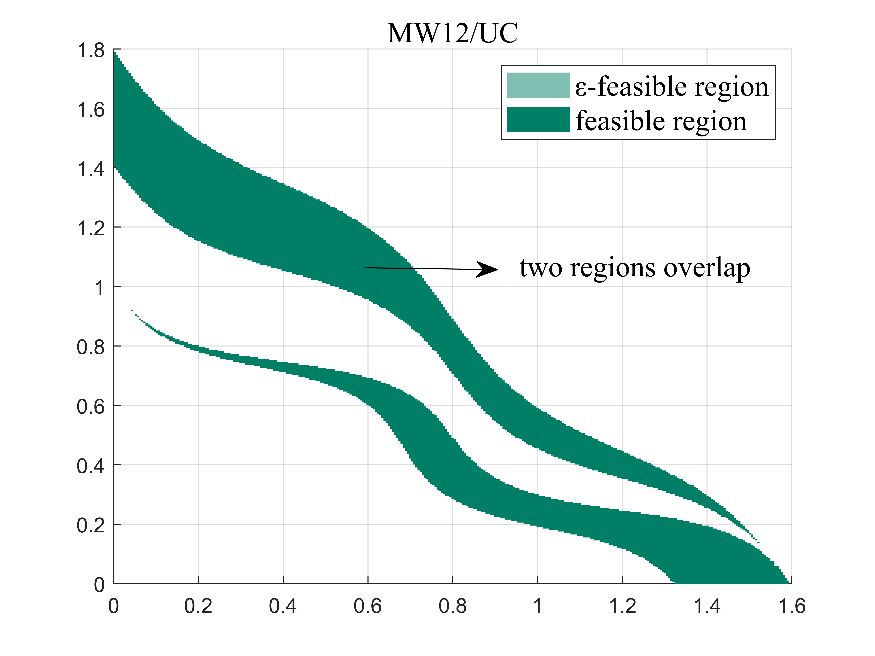}
		\text{(b)}
	\end{minipage}
	\centering
	\caption{Feasible region and $\varepsilon$-feasible region of (a) MW12 and (b)MW12/BC}
	\label{f2}
\end{figure}

Moreover, \(\varepsilon\)-based methods encounter significant challenges when addressing CMOP/BC. As shown in Fig. \ref{f2} (a), for MW12, the \(\varepsilon\)-based method treats solutions with CV values below a certain threshold as pseudo-feasible, thereby relaxing the original constraints and expanding the feasible region. However, in Fig. \ref{f2} (b) for the MW12/BC problem, the \(\varepsilon\)-based method loses its ability to effectively relax constraints, as the \(\varepsilon\)-feasible region completely coincides with the true feasible region.

This highlights the necessity of our work. We propose our algorithm to  address the challenges that current CMOEAs face when dealing with CMOP/BC problems.

\section{The Proposed Algorithm}
\label{s3}
This section provides a detailed overview of the implementation of the proposed DRMCMO. First, we introduce the basic principles of the Detection Region Method (DRM). Next, we explain the algorithmic framework of DRMCMO, followed by a brief analysis of DRM’s search behavior and its effectiveness.

\subsection{Detection Region Method}
In addressing CMOP/BC, the primary principle of DRM is to alleviate the stringent selection pressure on feasible solutions within the feasible region by relaxing the detection boundaries. Excessively high selection pressure can cause populations to become trapped in local feasible regions, making it difficult to escape. 

A solution \(\mathbf{x}\) is considered feasible in DRM if it satisfies at least one of the following conditions:
\begin{equation}
	\label{e8}
	\begin{cases}
		\textit{CV}(\mathbf{x}) = 0 \\
		\quad \quad \text{or} \\
		\varphi(\mathbf{x}) = \bigvee_{\textit{p} \in \textit{centers}} (||\mathbf{F}(\mathbf{x}) - \textit{p}|| < \textit{r})
	\end{cases}
\end{equation}
where \(\varphi(\mathbf{x})\) is a boolean function. The symbol \(\bigvee\) represents a logical "or" operation, indicating that if solution \(\mathbf{x}\) is within any detection region, then \(\varphi(\mathbf{x})\) returns true. The term \(\textit{centers}\) refers to the center points of all detection regions, \(\textit{p}\) denotes a specific center point, and \(\textit{r}\) is a dynamically changing threshold representing the radius of the detection region, calculated as follows:
\begin{equation}
	\label{e9}
	\textit{r} = (1 - \alpha) \cdot \textit{r}^\textit{max}
\end{equation}
where \(\textit{r}^\textit{max}\) is set to the modulus of the current population's target minimum vector after population initialization, and \(\alpha\) is a parameter controlling the change of \(\textit{r}\), calculated by:
\begin{equation}
	\label{e10}
	\alpha = \frac{1}{1 + \textit{e}^{-10\left(\frac{\textit{k} - \textit{k}_\textit{s}}{\textit{K} - \textit{k}_\textit{s}} - 0.6\right)}}
\end{equation}
where \(\textit{k}\) represents the current generation, \(\textit{k}_\textit{s}\) is the generation when DRM is activated, and \(\textit{K}\) is the maximum generation.

Generally, we aim for the parameter \(\alpha\) to increase from 0 to 1 over iterations, thereby controlling the size of the detection region from large to small. In this paper, we designed the S-shaped change in Equation \ref{e10}, which allows \(\alpha\) to increase gradually from 0, accelerate in the middle range, and then approach 1 slowly in the later stages. This pattern enables the detection region radius \(\textit{r}\) to initially shrink slowly, allowing the population to converge fully, then shrink rapidly in the middle stages, and finally shrink slowly again, guiding the population to explore near the CPF. We also experimented with other variations of \(\alpha\), including linear changes; however, our proposed S-shaped adjustment, as defined in Equation \ref{e10}, proved the most effective. For further details on the experimental design, please refer to the ``Ablation Study Results Analysis'' section below. The center of each detection region is determined based on the feasible solutions within the archive $\textit{Arch}$ and is also shifted to some extent based on 
$\textit{Arch}$. The set of centers can be expressed by the following formula:
\begin{equation}
		\label{e11}
	\textit{centers}=\{\textit{p}|\textit{p}=\mathbf{F(x)}+\alpha\cdot \textit{r}, \mathbf{x} \in \textit{Arch}\}
\end{equation}

Algorithm \ref{a1} outlines the DRM-based environment selection operation. In this algorithm, lines 1-5 detail the preprocessing of the merged population \textit{Q} using DRM. When the condition in equation (\ref{e8}) is satisfied, the constraint violation (CV) value of individuals in \textit{Q} is set to 0. This indicates that solutions within the detection regions are considered feasible in addition to those with \(\textit{CV} = 0\). The fitness calculation and environment selection operations in lines 6 and 7 follow the same procedures as those in SPEA2 \cite{zitzler2001spea2}. 

\begin{figure}[!htbp]
	\centering
	\includegraphics[width=6.2cm]{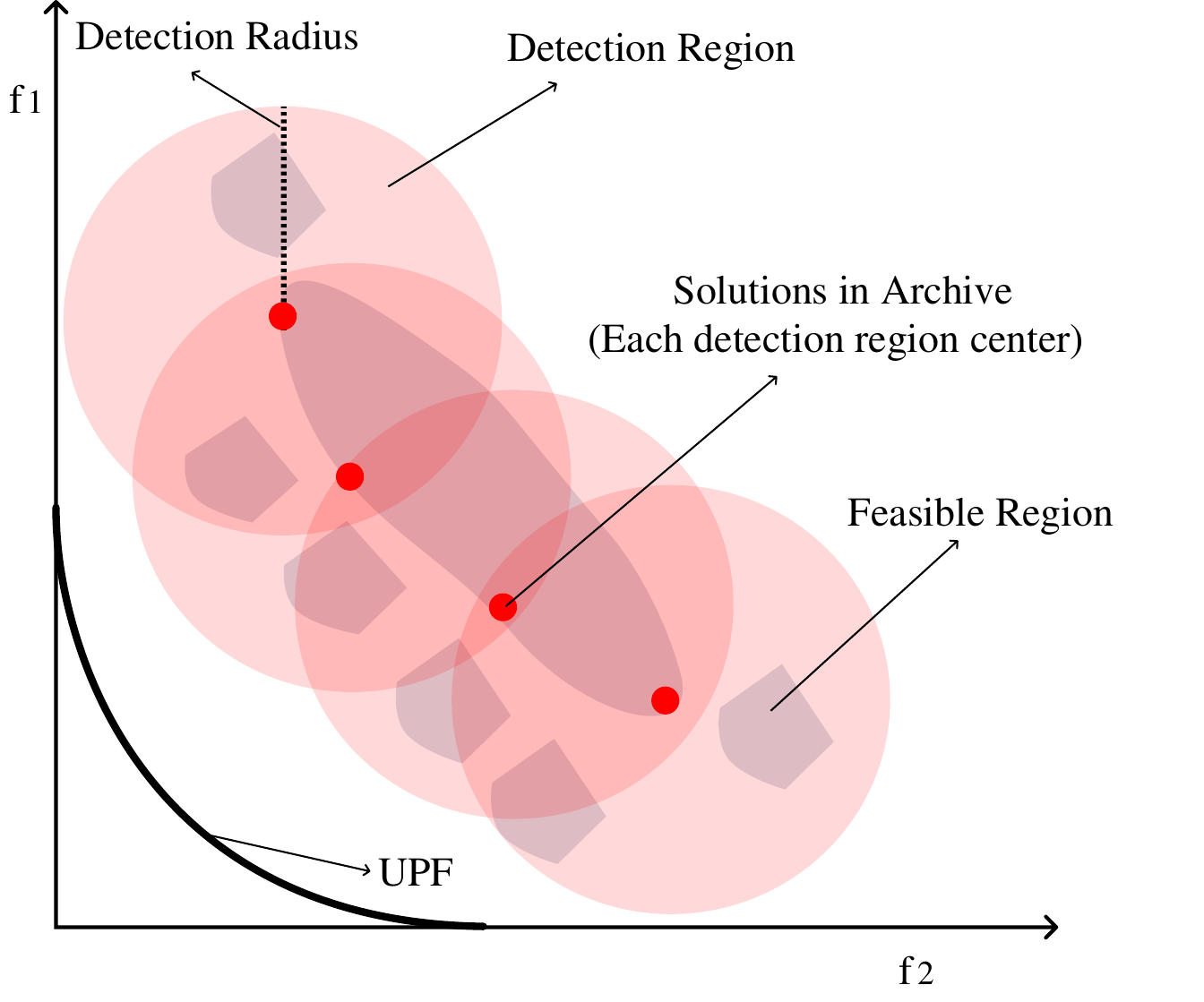}
	\centering
	
	\caption{An illustration of the early stage after the activation of DRM. The gray area represents the feasible region, and the red transparent regions represents all the detection regions.  }
	\label{f3}
\end{figure}

DRM replaces the \(\varepsilon\)-based method for relaxing constraints in CMOP/BC, effectively addressing its limitations. Figure \ref{f3} illustrates the initial activation of DRM, where the red circles are sufficiently large, allowing solutions that have locally converged due to excessive selection pressure to continue their convergence. The red points represent the centers of the detection regions, which correspond to feasible solutions within the archive (that have not yet shifted).

\begin{figure}[!htbp]
	\centering
	\includegraphics[width=6.2cm]{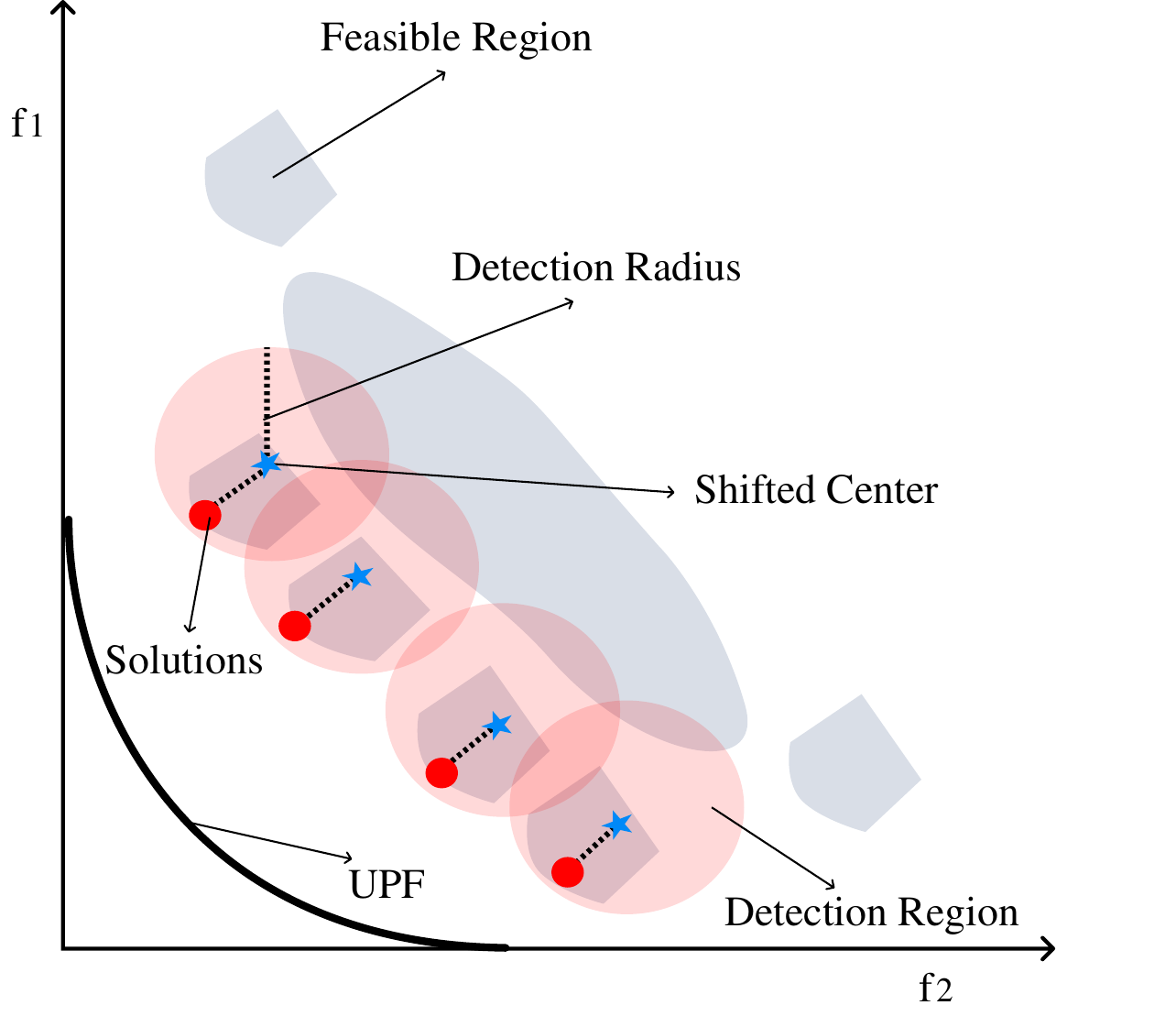}
	\centering
	
	\caption{An illustration of the later stage of the DRM. The red points are the solution set in the archive, and the blue points are the results after the red points are shifted, serving as the center points of the detection regions. The gray area represents the feasible region , and the red transparent region represents all the detection regions.  }
	\label{f4}
\end{figure}

As shown in Fig. \ref{f4}, as iterations progress, the detection regions shrink and the centers shift backward. This process prevents a significant accumulation of non-dominated solutions in the lower left corner of the feasible region, enabling the non-dominated set to contract to the actual feasible region before the detection regions are significantly reduced. Such an approach facilitates a more thorough exploration of the area near the feasible region, helping to identify additional feasible solutions and enhancing the diversity of the solution set. 

Moreover, once a new feasible solution is discovered, it serves as a new benchmark point for the detection regions, expanding their lateral search range. This mutual enhancement between feasible solutions and detection regions promotes a more effective search process.

\begin{algorithm}[!htb]
	\caption{environmental selection based on DRM} 
	\label{a1}
	\begin{algorithmic}[1]
		\REQUIRE \textit{N}: the population size.  \textit{Q}: combined population
		
		\ENSURE $\textit{Pop}_\textit{k}$ for the next generation;
		\FOR{$\mathbf{x}$ \textbf{in} \textit{Q}}
		\IF{$\mathbf{x}$ satisfies the conditions of Equation (\ref{e8})}
		\STATE $\textit{CV}(\mathbf{x})$ $\gets 0$: solutions in detection regions is also considered feasible solutions;
		\ENDIF 
		\ENDFOR
		\STATE \textit{fitness} $\gets$ Calculate fitness of $\textit{Q}$ based on CDP;
		\STATE $\textit{Pop}_\textit{k}$ $\gets$ Select the \textit{N} best solutions from \textit{Q} through environmental selection;
		\RETURN $\textit{Arch}$;

	\end{algorithmic}
\end{algorithm}

\subsection{Framework}
\begin{figure}[!htbp]
	\centering
	\includegraphics[width=8.5cm]{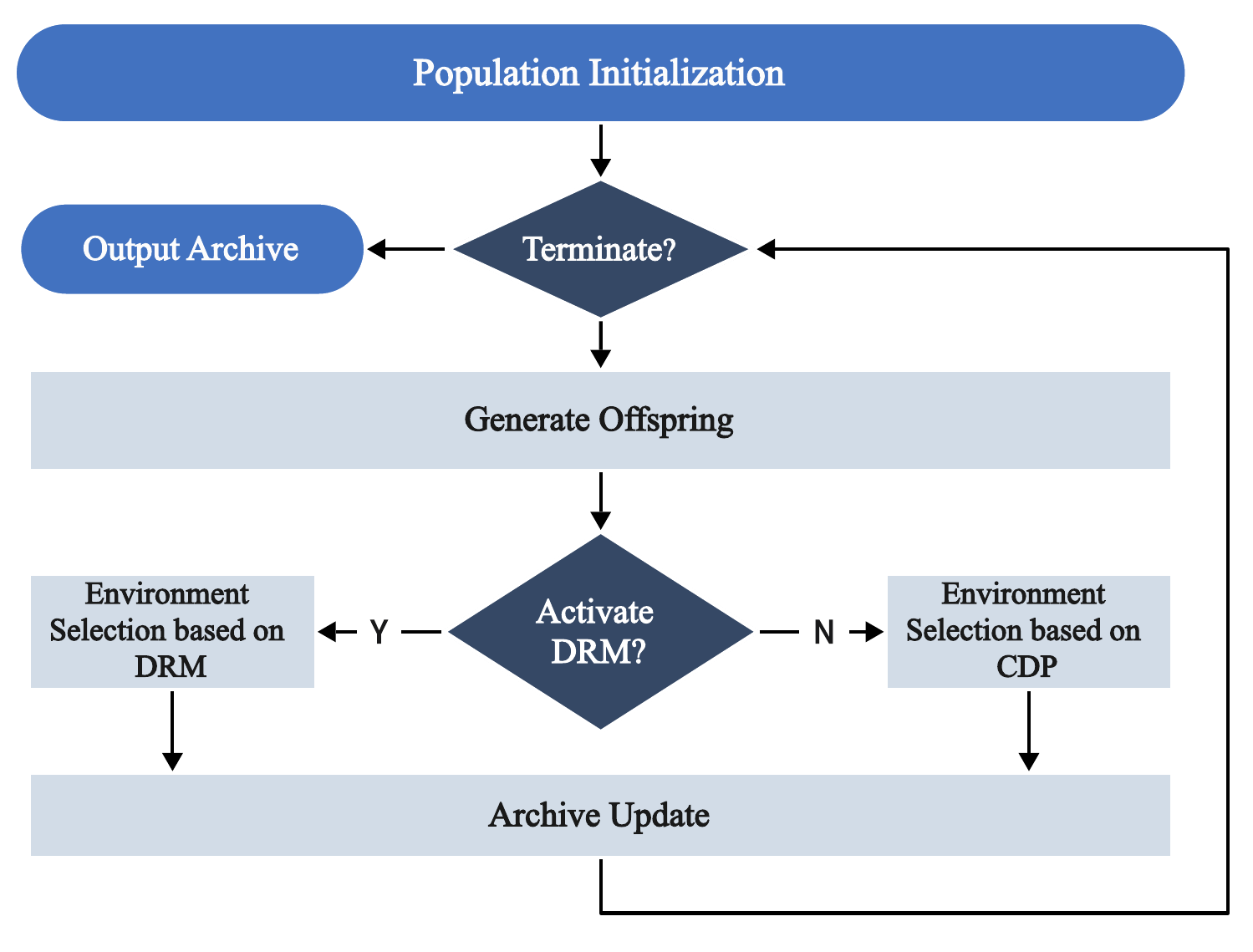}
	\centering
	
	\caption{The framework flowchart of the proposed DRMCMO.  }
	\label{f5}
\end{figure}

As a CHT, the DRM offers high portability and can be integrated with various algorithms. In this paper, we combine DRM with SPEA2 \cite{zitzler2001spea2} to create DRMCMO. SPEA2  was selected due to its proven competitiveness in numerous studies, particularly when addressing 2 to 3 objectives, where it demonstrates strong convergence and effective distribution of solutions.

Fig. \ref{f5} presents the overall framework of our proposed algorithm, which aligns with the conventional structure of evolutionary algorithms. An external archive set is utilized to store elite feasible solutions. The DRM is activated only after a feasible solution is found, as it requires a sufficient number of feasible solutions in the archive to establish a benchmark for subsequent searches.

\begin{algorithm}[!htb]
	\caption{The framework of DRMCMO} 
	\label{a2}
	\begin{algorithmic}[1]
		\REQUIRE \textit{N}: the population size.
		
		\ENSURE $\textit{Arch}$.
		\STATE $\textit{Pop}_0$ $\gets$ Generate the initial population of size $N$;
		\STATE \textit{Arch} (An external archive set) $\gets$  $\textit{Pop}_0$;
		\STATE  $\textit{r}^{\textit{max}}$ $\gets ||\mathbf{F}^{\text{min}} ||$;
		\STATE  \textit{k} $\gets 0$  (\textit{k} represents \textit{k}th generation);
		\STATE  $\textit{k}_\textit{s}$ $\gets 0$;
		\WHILE{the termination criterion is not satisfied}
		\STATE \textit{R} $\gets$ Generate offspring according to $\textit{Pop}_\textit{k}$;
		\STATE \textit{Q} $\gets$ \textit{R} $\cup $ $\textit{Pop}_\textit{k}$;
		\STATE \textit{k} $\gets$ \textit{k}+1;
		\IF { any feasible solution found in $\textit{Arch}$}
		\IF{ $\textit{k}_\textit{s} = 0$}
		\STATE  $\textit{k}_\textit{s}$ $\gets \textit{k}$;
		\ENDIF
		\STATE $\alpha$ $\gets$ update by Equation (\ref{e10});
		\STATE $\textit{r}$  $\gets$ update by Equation (\ref{e9});
		\STATE $\textit{Pop}_\textit{k}$ $\gets$ environmental selection for $\textit{Q}$ based on DRM (Algorithm \ref{a1});
		\ELSE
		\STATE $\textit{r}^{\textit{max}}$ $\gets ||\mathbf{F}^{\text{min}} ||$;
		\STATE \textit{fitness} $\gets$ Calculate fitness of $\textit{Q}$;
		\STATE $\textit{Pop}_\textit{k}$ $\gets$ environmental selection for $\textit{Q}$ based on CDP;
		\ENDIF
		
		\STATE \textit{Arch} $\gets$ UpdateArchive($\textit{Pop}_\textit{k}$\:$\cup$\:\textit{Arch}) based on CDP;
		
		\ENDWHILE
		\RETURN $\textit{Arch}$;

	\end{algorithmic}
\end{algorithm}

Algorithm 1 outlines the pseudocode for our framework. Lines 1 and 2 involve the initialization of the population and the archive, respectively. In line 3, \(\textit{r}^{\textit{max}}\) denotes the maximum radius of the detection regions, initially set to the magnitude of the minimal objective vector of the population. Line 4 indicates \textit{k}, the current generation, while line 5 records \(\textit{k}_\textit{s}\), the generation when DRM is activated. Lines 7-8 detail the reproduction and merging processes of the populations. 
Lines 18 to 20 highlight that in the early stages of evolution, if no feasible solutions exist in the archive, the search for feasible solutions primarily relies on the CDP for constraint handling. Although the guiding role of $\textit{CV}$ diminishes when addressing CMOP/BC, CDP continues to exert significant selective pressure on feasible solutions. The calculation of fitness and environment selection are basically consistent with SPEA2 \cite{zitzler2001spea2}. Given a solution $\mathbf{x}$ and a solution set $P$, the fitness can calculated as follow:
\begin{equation}
fitness(\mathbf{x})=\sum_{\mathbf{y}\in P,\mathbf{y} \prec \mathbf{x}}|R^{\mathbf{x}}|+\frac{1}{\sigma_{\mathbf{x}}^{k}+2},
\end{equation}
where $R^{\mathbf{x}}$ measures convergence and  represents the total number of solutions in the set $P$ that dominate $\mathbf{x}$ based on CDP. $\sigma_{\mathbf{x}}^{k}$ is the euclidean distance between  $\mathbf{x}$ its $k$-th neighbor. Meanwhile, \(\textit{r}^{\textit{max}}\) is continuously updated during the convergence process. Once a feasible solution is found in the archive, the algorithm switches strategies to our proposed DRM (lines 14 to 16) to reduce the selective pressure of CDP on feasible solutions, thereby preventing convergence to local optima. DRM is conceptually similar to \(\varepsilon\)-based methods, both aiming to dynamically control the relaxation of constraints to guide the population's search behavior. However, as discussed in Section \ref{s3} C, \(\varepsilon\)-based methods have critical shortcomings in dealing with CMOP/BC, which motivated the development of our DRM.

During the initial activation stage of the DRM, the detection range is broad, helping solutions trapped in local optima to escape and continue converging. This broad initial detection region also supports the neighbor pairing strategy, enhancing population diversity. Our study employs the neighbor pairing strategy from reference \cite{yuan2021indicator}. As iterations progress, the detection regions gradually contract, controlling the population's search behavior near the CPF. When new superior feasible solutions are discovered, the archive is updated (line 22), and these new feasible solutions establish a new detection region, further enhancing the population's diversity and distribution. Ultimately, as iterations continue, the detection regions contract within the feasible region, at which point the algorithm reverts to the search behavior based on CDP. The final output is the updated archive.

\begin{figure*}[!htbp]
	\centering
	\footnotesize
	\begin{minipage}{3.5cm}
		\centering
		\includegraphics[width=3.7cm]{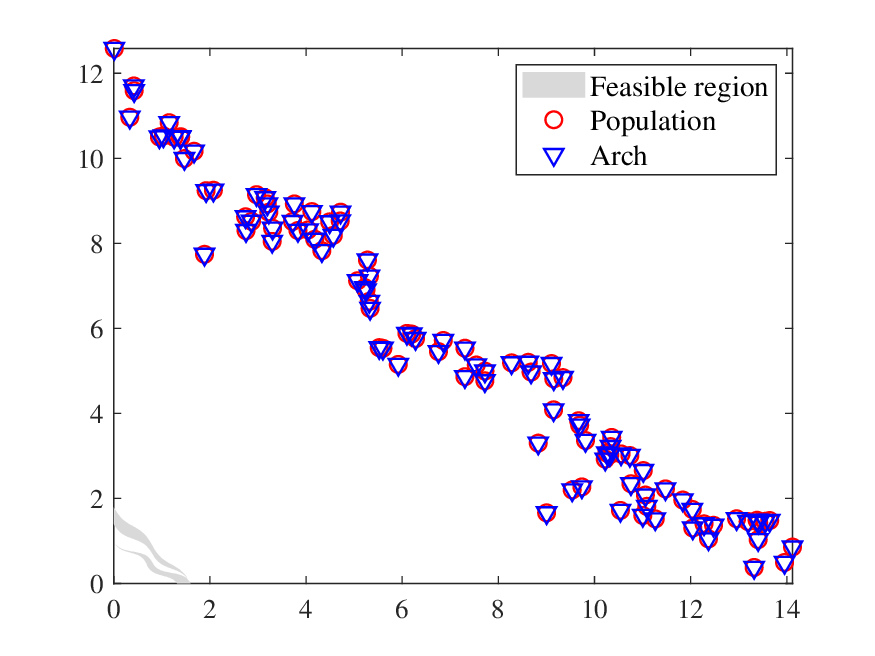}
		\text{(a)}
	\end{minipage}
	\begin{minipage}{3.5cm}
		\centering
		\includegraphics[width=3.7cm]{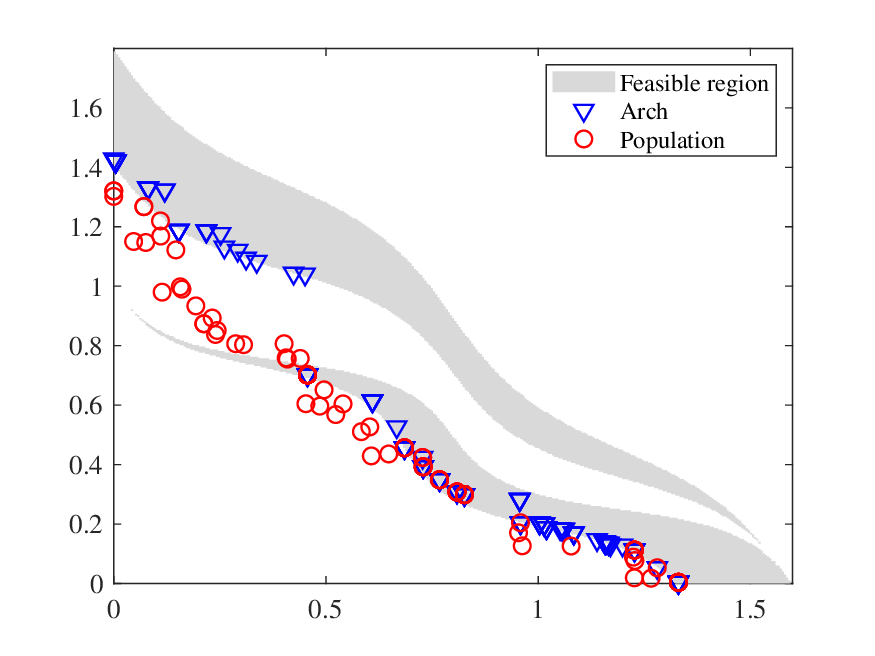}
		\text{(b)}
	\end{minipage}
	\begin{minipage}{3.5cm}
		\centering
		\includegraphics[width=3.7cm]{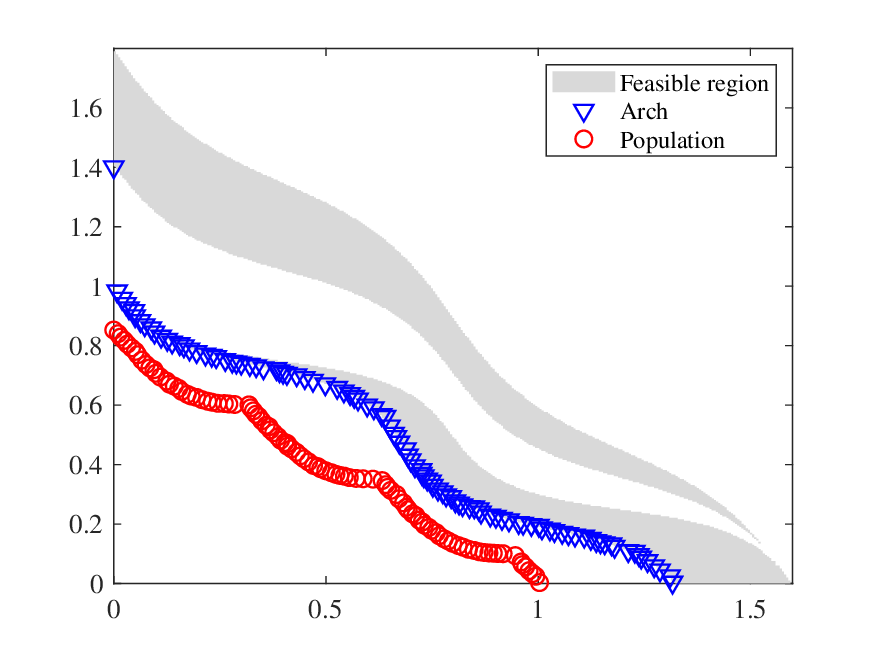}
		\text{(c)}
	\end{minipage}
	\begin{minipage}{3.5cm}
		\centering
		\includegraphics[width=3.7cm]{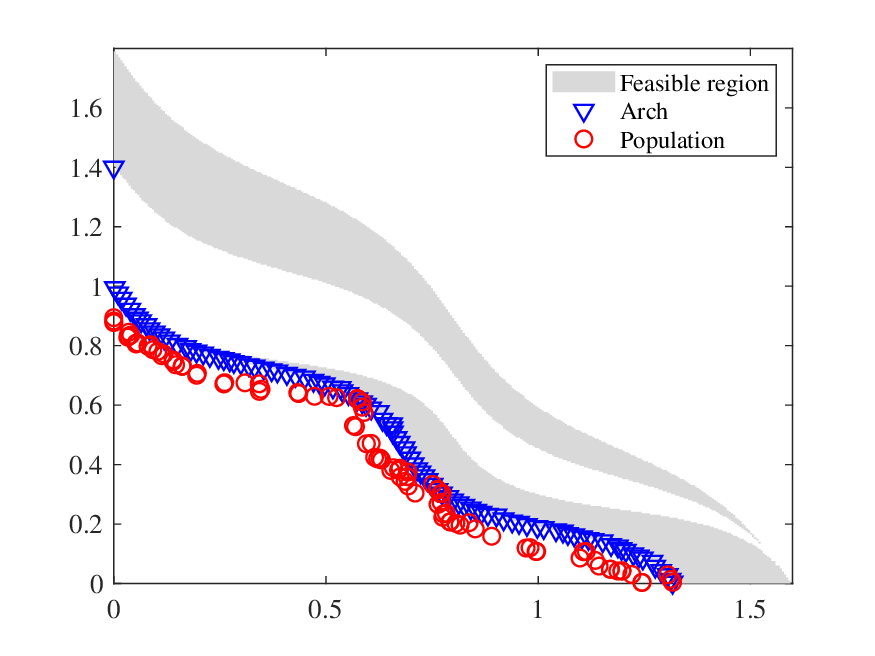}
		\text{(d)}
	\end{minipage}
	\begin{minipage}{3.5cm}
		\centering
		\includegraphics[width=3.7cm]{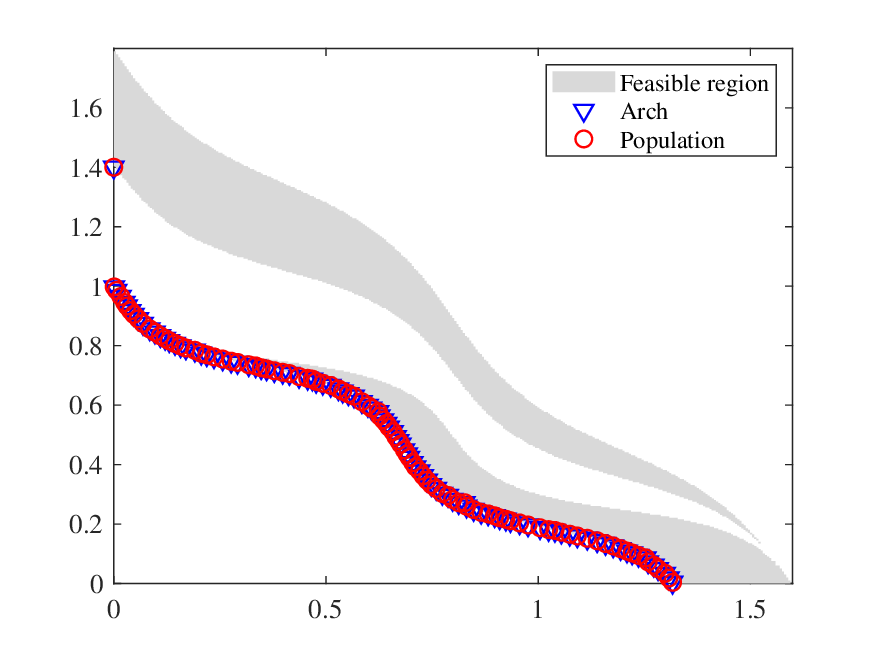}
		\text{(e)}
	\end{minipage}
	
	\caption{Search behavior of population and archive during 
		different  evolution process on MW12/BC. (a) First generation. (b)50th generation. (c) 200th generation.  (d) 600th generation. (e) 1000th(final) generation.}
	\label{f6}
\end{figure*}

To further analyze the search behavior of the proposed DRMCMO, we conducted preliminary experiments on MW12/BC. Fig. \ref{f6} presents the results of the population and the archive in DRMCMO across different generations. 
At the initial stage, the population is initialized, and the archive is initialized based on the population. By the 100th generation, some feasible solutions in the archive have settled into local optima and cannot converge further. However, with the activation of the detection regions, the population continues to converge due to the relaxation effect of these detection regions.  By the 200th generation, as the population continues to converge, it discovers  better feasible solutions. Consequently, the archive is updated, allowing it to escape from local optima. At this point, as shown in Fig. \ref{f6} (c), the population has converged to the UPF, and the archive effectively covers the CPF. By the 600th generation, as the detection regions continue to shrink, the population becomes concentrated near the true feasible region. This allows for the exploration of infeasible solutions close to the feasible region, potentially enhancing the diversity within the archive.  Finally, by the 1000th generation, the detection regions have fully contracted, leading to a consistent alignment between the population and the archive. The population has effectively explored the entire CPF, as illustrated in the Fig. \ref{f6} (e).
\subsection{Discussion}
In comparing DRMCMO with $\varepsilon$-constrained relaxation algorithms, the discussion focuses on how the DRM effectively overcomes the fundamental limitations of  $\varepsilon$-based methods when dealing with CMOP/BCs, thus demonstrating its unique advantages. $\varepsilon$-based constrained relaxation methods relax constraint violations by introducing a dynamic threshold, treating solutions with CV values below this threshold as "pseudo-feasible" solutions. This approach aims to expand the feasible region and alleviate excessive selection pressure towards feasible solutions. However, as the paper highlights, in CMOP/BC problems, the binary nature of the constraint functions (returning only 0 or 1) makes CV quantification inaccurate, and CV values are often identical across many regions. This renders the $\varepsilon$-based methods ineffective, as its "$\varepsilon$-feasible region" completely coincides with the true feasible region, failing to achieve effective constraint relaxation.

In contrast, DRM offers a novel and more adaptive approach. Instead of relying on inaccurate CV quantification, DRM relaxes constraints by establishing dynamic "detection regions" in the objective space around discovered feasible solutions. The radii of these regions are dynamically adjusted throughout the evolutionary process, starting with a larger radius to help solutions trapped in local optima escape. As the algorithm converges, the detection regions gradually shrink, guiding the population to explore areas closer to the true constrained Pareto front and enhancing solution diversity. Furthermore, whenever a new, high-quality feasible solution is found, DRM uses it as a new reference point to expand its lateral search scope, creating a mutual reinforcement mechanism between feasible solutions and detection regions that promotes a more effective search process.

Therefore, the advantage of DRMCMO lies in its robust framework, specifically designed to handle the challenges posed by CMOP/BCs. It relaxes constraints through a method independent of traditional CV quantification, making it more effective than Epsilon methods that rely on CV information. This approach not only mitigates selection pressure but also enables a more efficient exploration and convergence towards the true feasible region.

\subsection{Computational Complexity of DRMCMO}
The DRM in this paper is integrated into SPEA2 to form DRMCMO, whose computational complexity is primarily determined by three procedures: reproduction, fitness evaluation, and environmental selection. 

\begin{enumerate}[]
\item Reproduction: The time complexity for the neighbor pairing strategy is \(O(N)\), while the complexity for offspring generation is \(O(DN)\), where \(N\) represents the population size and \(D\) denotes the number of decision variables.

\item Fitness Evaluation: The time complexity for calculating the constraint violation (CV) and determining the dominance relationship is \(O((p + q)n)\) and \(O(mn^2)\), respectively. Here, \(M\) represents the number of objectives, \(p\) is the number of inequality constraints, and \(q\) is the number of equality constraints.

\item Environmental Selection: The time complexity for the DRM and truncation methods is \(O(mn^2)\) and \(O(n^3)\), respectively.
\end{enumerate}

In summary, the total computational complexity for each generation of DRMCMO is \(O((D + MN + p + q + N^2)n)\).

\section{Experimental Setup}
\label{s4}
In this section, we will detail the experimental setup, including the test problems used, comparative algorithms, evaluation metrics, and statistical testing methods, to ensure the rigor and reproducibility of this study. The following are the specific experimental settings.
\subsection{Benchmark Test Problems}
CMOP/BC is an emerging field that has yet to be extensively explored, with minimal related test problems and algorithms available. We drew inspiration from existing CMOP test problems to address this gap and designed a series of new test problems. Specifically, we adjusted the return values of the constraint functions for MW \cite{ma2019evolutionary}, LIRCMOP \cite{fan2019improved}, and DASCMOP \cite{fan2020difficulty}: when the constraint function value is greater than 0, it returns 1; when less than 0, it returns 0. These problems were renamed MW/BC, LIRCMOP/BC, and DASCMOP/BC. For equality constraints, we first converted them into inequality constraints and then binarized their return values. MW/BC suite includes 14 problems, inheriting various characteristics of actual CMOPs, such as smaller feasible regions, discontinuous CPF, and multiple nonlinear constraints. LIRCMOP/BC suite comprises 14 problems, primarily aimed at evaluating the performance of algorithms in dealing with large infeasible regions. DASCMOP/BC suite consists of 9 problems characterized by customizable constraint difficulty. Users can manually set parameters as needed to adjust the feasibility, convergence, and diversity levels of the problems.

These modified CMOP test problems retain the original characteristics and are perfectly adapted to the binary constraint nature of CMOP/BC, thereby facilitating the evaluation of algorithms in dealing with CMOP/BC. The number of evaluations for all problems is set to 100000. For parameter settings (objective number M and decision variable number D) of MW/BC, see Table \ref{t1}, and for LIRCMOP/BC and DASCMOP/BC, refer to Table \ref{t2}.

\subsection{Algorithm Settings}
Due to the scarcity of algorithms specifically designed for CMOP/BC, we selected several leading CMOEAs for comparative analysis. This comparison aims to highlight the limitations of existing CMOEAs in dealing with CMOP/BC challenges and demonstrate our proposed algorithm's superiority. Specifically, we selected two popular multi-stage algorithms, CMOEA-MS \cite{tian2021balancing} and MSCMO \cite{ma2021multi}, two multi-population algorithms, MCCMO \cite{zou2023multipopulation} and CDPEA \cite{ming2021dual}, and two multi-tasking-based algorithms, EMCMO \cite{qiao2022evolutionary} and IMTCMO \cite{qiao2023evolutionary}. These algorithms may utilize more than one CHTs. Notably, IMTCMO employs an $\varepsilon$-based method. It is noteworthy that all these algorithms utilize feasibility searches guided by CV.

All algorithms are configured according to the parameter settings proposed in their original publications and are independently run 30 times on each test problem. The population size of each algorithm was set to 100. Regarding the choice of search operators, IMTCMO utilizes a customized, modified DE operator \cite{qiao2023evolutionary}. The algorithm discussed in this paper, along with other comparative algorithms, was tested using both DE \cite{yu2013differential} and GA \cite{lambora2019genetic} operators, with the best-performing results from DE and GA being selected. The parameter settings for DE and GA are consistent with those described in reference \cite{tian2021balancing}.

\subsection{Performance Metric}

The Inverse Generational Distance (IGD) \cite{bosman2003balance} is a widely adopted metric for evaluating the performance of comparative algorithms Since it effectively reflects both the convergence and distribution of solutions. IGD can be defined as follows:
	\begin{eqnarray}\label{con:f5}
	I G D\left(P, P^{'}\right)=\frac{\sum_{\mathbf{v} \in P^{'}} d(\mathbf{v}, P)}{\left|P^{'}\right|},
	\end{eqnarray}
	where $P$ represents the set of solutions obtained by the algorithm, and $P^{'}$ is a set of reference points uniformly sampled along the constrained Pareto front. $d(\mathbf{v}, P)$ denotes the Euclidean distance between a solution $\mathbf{v}$ and its nearest neighbor in $P$.

	Hypervolume (HV) metric \cite{while2006faster} assesses the convergence and diversity of solutions by calculating the volume enclosed between the solution set produced by the algorithm and a predefined reference point. This metric is particularly valuable for evaluating algorithm performance on unknown problems, as it does not rely on knowledge of the true Pareto front, which is often unknown in real-world problems. A larger hypervolume indicates better performance. The definition of HV is as follows:
	\begin{eqnarray}\label{con:f6}
	HV(P)=\operatorname{V}\left(\bigcup_{x \in P}\left[f_{1}(x), z_{1}\right] \times \ldots\left[f_{m}(x), z_{m}\right]\right),
	\end{eqnarray}
	where $\operatorname{V}(\cdot)$ represents the Lebesgue measure; $m$ denotes the number of objectives and $\mathbf{z}=(z_{1}, \ldots, z_{m})^{T}$ is a reference point in the objective that can dominate the constraint PF.

In addition, the Wilcoxon rank-sum test \cite{zitzler2008quality} was employed in this study to analyze the results at a significance level of 0.05. The symbols “+”, “-”, and “$\approx$” were used to indicate whether the performance was significantly better, significantly worse, or statistically similar to the proposed algorithm, respectively. Furthermore, Friedman's test was utilized to compare the algorithms' performances across multiple datasets. The Friedman ranking provides a non-parametric method to assess multiple algorithms by ranking each algorithm's performance, where lower ranks indicate better performance. This allows for a comprehensive comparison of the algorithms' overall effectiveness.

\section{Experimental Results and Discussion}
\label{s5}
\subsection{Experimental Results on Benchmark Test Problems}
\begin{table*}[htbp!]
	\centering
	\setlength{\tabcolsep}{3pt}
	\caption{IGD  Results Obtained by the Proposed DRMCMO and All the Compared CMOEAs on  MW/BC Problems.}
	\begin{threeparttable}
	{\fontsize{6.58pt}{9.5pt}\selectfont
	\begin{tabular}{cccccccccc}
		\toprule
		\multirow{2}[4]{*}{Problem} & \multirow{2}[4]{*}{M} & \multicolumn{1}{c|}{\multirow{2}[4]{*}{D}} & \multicolumn{2}{c|}{Multi-Stage CMOEAs} & \multicolumn{2}{c|}{Multi-Population CMOEAs} & \multicolumn{2}{c|}{Multi-Task CMOEAs} & Proposed CMOEA \\
		\cmidrule{4-10}          &       & \multicolumn{1}{c|}{} & CMOEA\_MS & \multicolumn{1}{c|}{MSCMO} & cDPEA & \multicolumn{1}{c|}{MCCMO} & EMCMO & \multicolumn{1}{c|}{IMTCMO} & DRMCMO \\
		\midrule
		MW/BC1 & 2     & \multicolumn{1}{c|}{15} & 2.3801e-3 (1.90e-3) - & \multicolumn{1}{c|}{1.6161e-3 (1.49e-5) $\approx$} & 1.6177e-3 (1.53e-5) $\approx$ & \multicolumn{1}{c|}{1.6177e-3 (9.78e-6) $\approx$} & \hl{1.5997e-3 (1.01e-5) $\approx$} & \multicolumn{1}{c|}{1.6578e-3 (2.16e-5) $\approx$} & 1.6853e-3 (1.38e-5)  \\
		MW/BC2 & 2     & \multicolumn{1}{c|}{15} & 2.4363e-2 (7.09e-3) - & \multicolumn{1}{c|}{2.3138e-2 (9.49e-3) -} & 1.9801e-2 (8.31e-3) - & \multicolumn{1}{c|}{2.1840e-2 (1.27e-2) -} & 1.4601e-2 (7.96e-3) - & \multicolumn{1}{c|}{5.4816e-2 (2.03e-2) -} & \hl{4.9427e-3 (4.06e-4) }\\
		MW/BC3 & 2     & \multicolumn{1}{c|}{15} & 5.3578e-3 (4.09e-4) - & \multicolumn{1}{c|}{5.0740e-3 (4.13e-4) $\approx$} & 4.8840e-3 (3.09e-4) $\approx$ & \multicolumn{1}{c|}{4.8811e-3 (2.46e-4) $\approx$} & 4.8921e-3 (2.02e-4) $\approx$ & \multicolumn{1}{c|}{4.8788e-3 (1.20e-4) $\approx$} & \hl{4.8440e-3 (1.74e-4)  }\\
		MW/BC4 & 2     & \multicolumn{1}{c|}{15} & 4.9348e-3 (2.13e-4) + & \multicolumn{1}{c|}{5.0056e-3 (5.42e-4) $\approx$} & 4.8046e-3 (3.42e-4) + & \multicolumn{1}{c|}{4.8926e-3 (3.59e-4) +} & \hl{4.6388e-3 (4.15e-4) + }& \multicolumn{1}{c|}{7.0803e-3 (2.29e-4) $\approx$} & 5.0788e-3 (2.36e-4) \\
		MW/BC5 & 2     & \multicolumn{1}{c|}{15} & 1.9557e-2 (1.14e-2) - & \multicolumn{1}{c|}{1.6369e-2 (4.49e-2) -} & \hl{5.7527e-4 (5.11e-4) + }& \multicolumn{1}{c|}{1.6302e-3 (8.55e-4) -} & 1.6505e-2 (5.38e-2) - & \multicolumn{1}{c|}{2.7214e-3 (5.01e-4) -} & 1.3136e-3 (4.00e-4) \\
		MW/BC6 & 2     & \multicolumn{1}{c|}{15} & 2.3168e-2 (1.13e-2) - & \multicolumn{1}{c|}{1.8057e-2 (6.84e-3) -} & 1.6433e-2 (9.00e-3) - & \multicolumn{1}{c|}{1.3706e-2 (7.41e-3) -} & 1.2989e-2 (5.67e-3) - & \multicolumn{1}{c|}{1.8871e-1 (1.95e-1) -} & \hl{2.7871e-3 (2.28e-5) }\\
		MW/BC7 & 2     & \multicolumn{1}{c|}{15} & 2.3685e-2 (2.42e-2) - & \multicolumn{1}{c|}{4.5717e-3 (2.73e-4) -} & 4.3410e-3 (2.85e-4) - & \multicolumn{1}{c|}{4.5882e-3 (3.61e-4) -} & 4.9278e-3 (4.93e-4) - & \multicolumn{1}{c|}{4.2648e-3 (1.75e-4) $\approx$} & \hl{4.2563e-3 (1.92e-4)  }\\
		MW/BC8 & 2     & \multicolumn{1}{c|}{15} & 1.8288e-2 (8.11e-3) - & \multicolumn{1}{c|}{2.9295e-2 (1.55e-2) -} & 1.3200e-2 (1.04e-2) - & \multicolumn{1}{c|}{2.1195e-2 (1.21e-2) -} & 7.9340e-3 (4.81e-3) - & \multicolumn{1}{c|}{4.2156e-2 (1.80e-2) -} & \hl{2.1640e-3 (2.59e-5) }\\
		MW/BC9 & 2     & \multicolumn{1}{c|}{15} & 8.4447e-2 (2.03e-1) - & \multicolumn{1}{c|}{5.0470e-3 (1.66e-3) $\approx$} & 6.3542e-2 (2.04e-1) - & \multicolumn{1}{c|}{6.3201e-2 (2.04e-1) $\approx$} & 4.7537e-3 (2.20e-4) - & \multicolumn{1}{c|}{1.5559e-2 (3.99e-3) -} & \hl{4.4964e-3 (2.73e-4) }\\
		MW/BC10 & 2     & \multicolumn{1}{c|}{15} & 7.2195e-2 (5.85e-2) - & \multicolumn{1}{c|}{4.4035e-2 (3.13e-2) -} & 2.6225e-2 (1.62e-2) - & \multicolumn{1}{c|}{4.5549e-2 (5.14e-2) -} & 2.5395e-2 (2.38e-2) - & \multicolumn{1}{c|}{2.4347e-1 (2.36e-1) -} & \hl{3.6211e-3 (5.01e-5) }\\
		MW/BC11 & 2     & \multicolumn{1}{c|}{15} & 6.2411e-3 (2.31e-4) $\approx$ & \multicolumn{1}{c|}{9.5219e-3 (4.83e-3) -} & 6.1584e-3 (1.33e-4) $\approx$ & \multicolumn{1}{c|}{\hl{5.8911e-3 (1.32e-4) +}} & 6.0720e-3 (2.99e-4) $\approx$ & \multicolumn{1}{c|}{5.9249e-3 (1.16e-4) $\approx$} & 6.0765e-3 (1.34e-4) \\
		MW/BC12 & 2     & \multicolumn{1}{c|}{15} & 5.2660e-3 (3.15e-4) - & \multicolumn{1}{c|}{5.9823e-2 (1.91e-1) -} & 4.8327e-3 (6.68e-5) $\approx$ & \multicolumn{1}{c|}{6.9639e-3 (8.07e-3) -} & 1.0031e-2 (1.80e-2) - & \multicolumn{1}{c|}{4.8132e-3 (1.16e-4) $\approx$} & \hl{4.7391e-3 (9.79e-5) }\\
		MW/BC13 & 2     & \multicolumn{1}{c|}{15} & 1.0837e-1 (5.11e-2) - & \multicolumn{1}{c|}{9.4449e-2 (3.09e-2) -} & 4.3173e-2 (2.91e-2) - & \multicolumn{1}{c|}{5.1890e-2 (4.08e-2) -} & 2.9448e-2 (2.54e-2) - & \multicolumn{1}{c|}{3.4548e-1 (4.32e-1) -} & \hl{1.3118e-2 (3.37e-3) }\\
		MW/BC14 & 2     & \multicolumn{1}{c|}{15} & 3.3393e-2 (2.22e-3) - & \multicolumn{1}{c|}{1.6685e-2 (1.03e-3) -} & 1.5519e-2 (8.33e-4) $\approx$ & \multicolumn{1}{c|}{1.6955e-2 (1.94e-3) -} & 1.5897e-2 (2.49e-3) $\approx$ & \multicolumn{1}{c|}{1.6104e-2 (5.65e-4) -} & \hl{1.5283e-2 (5.18e-4) }\\
		\midrule
		\multicolumn{3}{c}{+/-/$\approx$} & 1/12/1 & 0/10/4 & 2/7/5 & 2/9/3 & 1/9/4 & 0/8/6 &  \\
		\bottomrule
	\end{tabular}%
	}
		\begin{tablenotes}
		\item The best result for each row is highlighted.``N/A'' means that no feasible solution can be found. ``+'', ``-'' and ``$\approx$'' indicate that the result is significantly \\better, significantly worse, and  statistically similar to the results obtained by DRMCMO,respectively.
	\end{tablenotes}
	\end{threeparttable}
	\label{t1}%
\end{table*}%

\begin{figure}[!htbp]
	\footnotesize
	\centering
	\includegraphics[width=9.2cm]{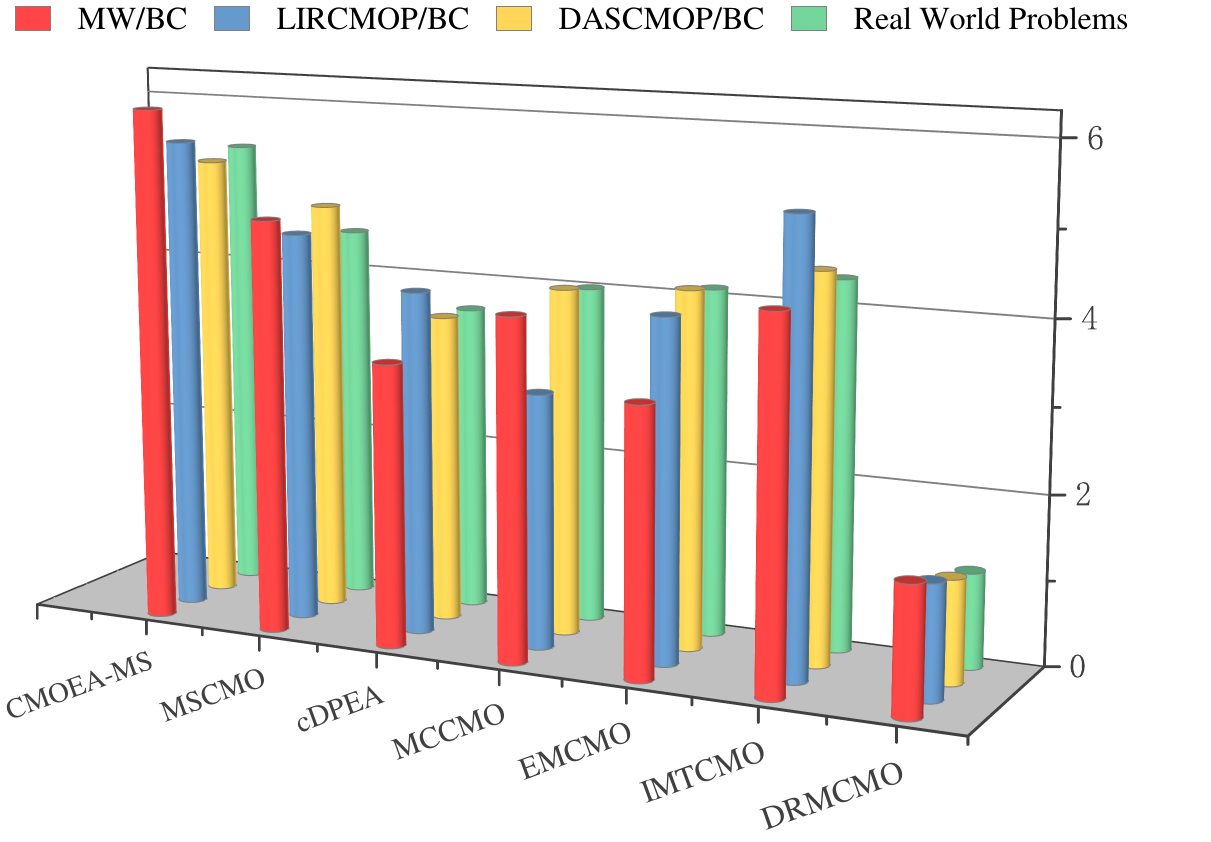}
	\centering
	
	\caption{ Friedman’s test among the seven algorithms concerning IGD on MW/BC, LIRCMOP/BC, DASCMOP/BC and real world problems.}
	\label{f7}
\end{figure}

The MW test suite encompasses a diverse range of complex constrained scenarios, including small feasible regions, dispersed feasible regions, nonlinear constraints, and various shapes of CPF. Building upon the foundation of the original MW problems,  the MW/BC introduces even more challenges.   All constraints in this version do not provide precise, quantifiable feedback; they only provide binary outcomes. Experimental results have shown that this modification presents severe challenges to existing state-of-the-art CMOEAs. Table \ref{t1} illustrates the average IGD and standard deviation obtained by the proposed DRMCMO and other compared CMOEAs over 30 independent runs on the MW/BC test suite. DRMCMO achieved the best IGD on 22 questions. Compared with CMOEA-MS, MSCMO, MCCMO, cPDEA, EMCMO, and IMTCMO, DRMCMO is significantly better in 12, 10, 7, 9, and 9 problems, respectively. We conducted a Friedman test on the MW/BC suite. As displayed in Fig. \ref{f7}, the results of the average friedman rankings for all algorithms indicate that DRMCMO leads with a ranking of 1.51, significantly outperforming its peers.

 \begin{figure}[!htbp]
	\centering
	\footnotesize
	\begin{minipage}{4.2cm}
		\centering
		\includegraphics[width=4.2cm]{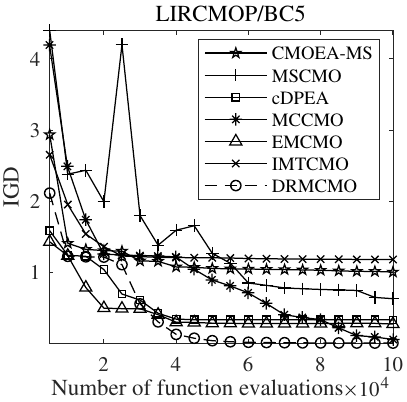}
		\text{(a)}
	\end{minipage}
	\begin{minipage}{4.4cm}
		\centering
		\includegraphics[width=4.4cm]{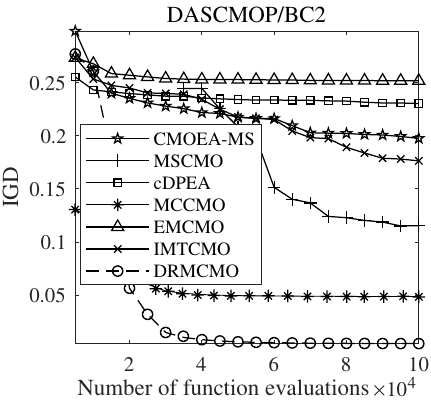}
		\text{(b)}
	\end{minipage}
	\centering
	\caption{IGD curves from 30 independent runs of all algorithms on (a) LIRCMOP/BC5 and (b) DASCMOP/BC2.}
	\label{IGDC}
\end{figure}

\begin{table*}[htbp]
	\centering
	\setlength{\tabcolsep}{3pt}
	\caption{IGD  Results Obtained by the Proposed DRMCMO and All the Compared CMOEAs on LIRCMOP/BC and DASCMOP/BC.}
	\begin{threeparttable}
		{\fontsize{6.58pt}{9.5pt}\selectfont
			\begin{tabular}{cccccccccc}
				\toprule
				\multirow{2}[4]{*}{Problem} & \multirow{2}[4]{*}{M} & \multicolumn{1}{c|}{\multirow{2}[4]{*}{D}} & \multicolumn{2}{c|}{Multi-Stage CMOEAs} & \multicolumn{2}{c|}{Multi-Population CMOEAs} & \multicolumn{2}{c|}{Multi-Task CMOEAs} & Proposed CMOEA \\
				\cmidrule{4-10}          &       & \multicolumn{1}{c|}{} & CMOEA\_MS & \multicolumn{1}{c|}{MSCMO} & cDPEA & \multicolumn{1}{c|}{MCCMO} & EMCMO & \multicolumn{1}{c|}{IMTCMO} & DRMCMO \\
				\midrule
				LIRCMOP/BC1 & 2     & \multicolumn{1}{c|}{30} & 3.1128e-1 (1.35e-1) - & \multicolumn{1}{c|}{3.5812e-1 (4.73e-2) -} & 3.2054e-1 (8.03e-2) - & \multicolumn{1}{c|}{2.3267e-1 (1.29e-1) $\approx$} & 2.5289e-1 (5.49e-2) - & \multicolumn{1}{c|}{2.3088e-1 (6.14e-2) -} & \hl{1.6979e-1 (7.62e-2) }\\
				LIRCMOP/BC2 & 2     & \multicolumn{1}{c|}{30} & 1.9293e-1 (8.29e-2) - & \multicolumn{1}{c|}{6.8282e-1 (0.00e+0) $\approx$} & 2.3895e-1 (5.02e-2) - & \multicolumn{1}{c|}{1.1634e-1 (7.84e-2) $\approx$} & 2.5815e-1 (3.90e-2) - & \multicolumn{1}{c|}{2.0900e-1 (3.80e-2) -} & \hl{1.0770e-1 (4.37e-2) }\\
				LIRCMOP/BC3 & 2     & \multicolumn{1}{c|}{30} & 2.3082e-1 (1.10e-1) $\approx$ & \multicolumn{1}{c|}{NaN (NaN)} & 2.8805e-1 (1.70e-1) $\approx$ & \multicolumn{1}{c|}{3.3303e-1 (9.70e-2) -} & 2.9102e-1 (6.50e-2) - & \multicolumn{1}{c|}{\hl{1.6544e-1 (6.29e-2) $\approx$}} & 2.0470e-1 (7.64e-2) \\
				LIRCMOP/BC4 & 2     & \multicolumn{1}{c|}{30} & 2.4998e-1 (8.73e-2) - & \multicolumn{1}{c|}{1.7371e-1 (4.83e-2) $\approx$} & 2.4092e-1 (6.21e-2) - & \multicolumn{1}{c|}{3.0013e-1 (1.21e-1) -} & 2.7123e-1 (4.89e-2) - & \multicolumn{1}{c|}{1.7726e-1 (5.85e-2) $\approx$} & \hl{1.5727e-1 (2.78e-2)  }\\
				LIRCMOP/BC5 & 2     & \multicolumn{1}{c|}{30} & 1.0061e+0 (4.39e-1) - & \multicolumn{1}{c|}{6.3171e-1 (6.15e-1) -} & 3.3413e-1 (5.96e-2) - & \multicolumn{1}{c|}{5.8861e-2 (7.73e-2) -} & 2.7554e-1 (6.85e-2) - & \multicolumn{1}{c|}{1.1823e+0 (2.56e-2) -} & \hl{8.1548e-3 (5.30e-4) }\\
				LIRCMOP/BC6 & 2     & \multicolumn{1}{c|}{30} & 1.2052e+0 (3.32e-1) - & \multicolumn{1}{c|}{5.9791e-1 (5.79e-1) -} & 3.2788e-1 (9.58e-2) - & \multicolumn{1}{c|}{5.0242e-2 (7.37e-2) -} & 3.3299e-1 (6.08e-2) - & \multicolumn{1}{c|}{1.2657e+0 (2.76e-1) -} & \hl{7.5513e-3 (6.57e-4) }\\
				LIRCMOP/BC7 & 2     & \multicolumn{1}{c|}{30} & 1.4158e+0 (6.21e-1) - & \multicolumn{1}{c|}{1.8039e-1 (3.86e-1) -} & 1.3064e-1 (2.52e-2) - & \multicolumn{1}{c|}{2.0963e-2 (3.39e-2) -} & 1.2021e-1 (3.57e-2) - & \multicolumn{1}{c|}{6.5947e-1 (7.59e-1) -} & \hl{8.3228e-3 (7.10e-4) }\\
				LIRCMOP/BC8 & 2     & \multicolumn{1}{c|}{30} & 1.1120e+0 (7.19e-1) - & \multicolumn{1}{c|}{2.4965e-1 (4.92e-1) -} & 1.7897e-1 (5.20e-2) - & \multicolumn{1}{c|}{8.5531e-3 (5.29e-4) $\approx$} & 1.9878e-1 (5.23e-2) - & \multicolumn{1}{c|}{1.1221e+0 (7.07e-1) -} & \hl{8.4891e-3 (9.45e-4) }\\
				LIRCMOP/BC9 & 2     & \multicolumn{1}{c|}{30} & 5.3652e-1 (6.49e-2) - & \multicolumn{1}{c|}{3.3507e-1 (2.11e-1) -} & 4.4726e-1 (1.19e-1) - & \multicolumn{1}{c|}{1.1951e-1 (4.29e-2) -} & 5.4026e-1 (1.31e-1) - & \multicolumn{1}{c|}{4.9798e-1 (9.48e-2) -} & \hl{5.6304e-2 (3.38e-2) }\\
				LIRCMOP/BC10 & 2     & \multicolumn{1}{c|}{30} & 3.9755e-1 (5.24e-2) - & \multicolumn{1}{c|}{1.9426e-1 (1.22e-1) -} & 2.7913e-1 (9.78e-2) - & \multicolumn{1}{c|}{1.4958e-2 (3.07e-3) -} & 1.8352e-1 (8.18e-2) - & \multicolumn{1}{c|}{2.5042e-1 (1.17e-1) -} & \hl{7.0248e-3 (8.71e-4) }\\
				LIRCMOP/BC11 & 2     & \multicolumn{1}{c|}{30} & 4.0405e-1 (8.77e-2) - & \multicolumn{1}{c|}{8.9515e-2 (1.13e-1) -} & 1.0209e-1 (4.63e-2) - & \multicolumn{1}{c|}{1.1021e-2 (7.31e-3) -} & 8.1008e-2 (5.30e-2) - & \multicolumn{1}{c|}{4.6341e-1 (1.61e-1) -} & \hl{2.6527e-3 (1.51e-4) }\\
				LIRCMOP/BC12 & 2     & \multicolumn{1}{c|}{30} & 2.8821e-1 (4.90e-2) - & \multicolumn{1}{c|}{1.1639e-1 (9.11e-2) -} & 1.8035e-1 (8.19e-2) - & \multicolumn{1}{c|}{2.9664e-2 (4.59e-2) -} & 2.3924e-1 (6.90e-2) - & \multicolumn{1}{c|}{3.2463e-1 (8.37e-2) -} & \hl{3.2715e-3 (3.39e-4) }\\
				LIRCMOP/BC13 & 3     & \multicolumn{1}{c|}{30} & 6.5785e-1 (3.48e-1) - & \multicolumn{1}{c|}{1.4207e-1 (3.73e-2) -} & 9.3233e-2 (9.65e-4) + & \multicolumn{1}{c|}{1.2085e-1 (2.02e-3) -} & \hl{9.1240e-2 (9.33e-4) +} & \multicolumn{1}{c|}{1.3107e+0 (1.20e-3) -} & 1.1655e-1 (2.54e-3) \\
				LIRCMOP/BC14 & 3     & \multicolumn{1}{c|}{30} & 6.6212e-1 (4.31e-1) - & \multicolumn{1}{c|}{1.7376e-1 (4.30e-2) -} & 9.5442e-2 (6.98e-4) + & \multicolumn{1}{c|}{1.0432e-1 (1.36e-3) -} & \hl{9.4537e-2 (7.87e-4) +} & \multicolumn{1}{c|}{1.2465e+0 (7.20e-2) -} & 1.0043e-1 (1.93e-3) \\
				\midrule
				DASCMOP/BC1 & 2     & \multicolumn{1}{c|}{30} & 6.0967e-1 (2.13e-1) - & \multicolumn{1}{c|}{2.2897e-1 (3.27e-1) -} & 6.6952e-1 (4.73e-2) - & \multicolumn{1}{c|}{4.2892e-3 (1.82e-3) -} & 7.2308e-1 (4.33e-2) - & \multicolumn{1}{c|}{4.1764e-1 (3.67e-1) -} & \hl{4.1491e-3 (6.42e-4) }\\
				DASCMOP/BC2 & 2     & \multicolumn{1}{c|}{30} & 1.9762e-1 (7.24e-2) - & \multicolumn{1}{c|}{1.1540e-1 (1.01e-1) -} & 2.3049e-1 (2.45e-2) - & \multicolumn{1}{c|}{4.9839e-2 (1.77e-4) -} & 2.5182e-1 (2.43e-2) - & \multicolumn{1}{c|}{1.7643e-1 (1.10e-1) -} & \hl{4.3764e-3 (7.17e-5) }\\
				DASCMOP/BC3 & 2     & \multicolumn{1}{c|}{30} & 3.4442e-1 (6.73e-3) - & \multicolumn{1}{c|}{NaN (NaN)} & 3.6258e-1 (5.50e-2) - & \multicolumn{1}{c|}{1.9744e-1 (1.14e-1) -} & 3.2365e-1 (5.97e-2) - & \multicolumn{1}{c|}{3.4641e-1 (5.57e-2) -} & \hl{1.7340e-2 (3.21e-3) }\\
				DASCMOP/BC4 & 2     & \multicolumn{1}{c|}{30} & 1.2044e-1 (1.31e-1) - & \multicolumn{1}{c|}{1.4020e-2 (7.54e-3) -} & 3.3739e-3 (4.30e-3) - & \multicolumn{1}{c|}{1.5525e-1 (2.71e-1) -} & 2.0069e-3 (1.31e-3) - & \multicolumn{1}{c|}{1.9998e-1 (1.24e-1) -} & \hl{1.3868e-3 (1.38e-4) }\\
				DASCMOP/BC5 & 2     & \multicolumn{1}{c|}{30} & 1.3202e-1 (1.12e-1) - & \multicolumn{1}{c|}{5.1552e-2 (3.28e-2) -} & \hl{2.7981e-3 (8.29e-5) +} & \multicolumn{1}{c|}{1.5392e-1 (2.56e-1) -} & 2.9299e-3 (1.70e-4) $\approx$ & \multicolumn{1}{c|}{1.5585e-1 (1.83e-1) -} & 2.9065e-3 (1.26e-4) \\
				DASCMOP/BC6 & 2     & \multicolumn{1}{c|}{30} & 2.1223e-1 (1.76e-1) - & \multicolumn{1}{c|}{2.8890e-1 (2.73e-1) -} & 2.5832e-2 (2.17e-2) - & \multicolumn{1}{c|}{1.3289e-1 (2.77e-1) -} & 2.5589e-2 (9.88e-3) - & \multicolumn{1}{c|}{4.3298e-1 (1.58e-1) -} & \hl{1.9786e-2 (1.76e-3) }\\
				DASCMOP/BC7 & 3     & \multicolumn{1}{c|}{30} & 2.1764e-1 (5.73e-2) - & \multicolumn{1}{c|}{2.9194e-1 (3.34e-1) -} & 3.1277e-2 (9.37e-4) - & \multicolumn{1}{c|}{3.2942e-1 (4.56e-1) -} & 3.2899e-2 (1.20e-3) - & \multicolumn{1}{c|}{3.1991e-2 (2.40e-3) -} & \hl{3.0213e-2 (3.78e-4)  }\\
				DASCMOP/BC8 & 3     & \multicolumn{1}{c|}{30} & 3.5075e-1 (2.57e-1) - & \multicolumn{1}{c|}{2.9233e-1 (2.08e-1) -} & 4.0815e-2 (1.69e-3) $\approx$ & \multicolumn{1}{c|}{1.1397e-1 (1.24e-1) -} & 4.1173e-2 (1.95e-3) $\approx$ & \multicolumn{1}{c|}{4.3028e-2 (5.01e-3) -} & \hl{3.9592e-2 (8.30e-4)  }\\
				DASCMOP/BC9 & 3     & \multicolumn{1}{c|}{30} & 2.7156e-1 (1.47e-1) - & \multicolumn{1}{c|}{8.1463e-2 (1.35e-2) -} & 2.1155e-1 (4.69e-2) - & \multicolumn{1}{c|}{6.3909e-2 (7.64e-2) -} & 3.2514e-1 (4.78e-2) - & \multicolumn{1}{c|}{\hl{3.9814e-2 (1.01e-3) +}} & 4.1146e-2 (1.59e-3) \\
				\midrule
				\multicolumn{3}{c}{+/-/$\approx$} & 0/22/1 & 0/19/2 & 3/18/2 & 0/20/3 & 2/19/2 & 1/20/2 &  \\
				\bottomrule
			\end{tabular}%
		}
		\begin{tablenotes}
			\item The best result for each row is highlighted.``N/A'' means that no feasible solution can be found. ``+'', ``-'' and ``$\approx$'' indicate that the result is significantly \\better, significantly worse, and  statistically similar to the results obtained by DRMCMO,respectively.
		\end{tablenotes}
	\end{threeparttable}
	\label{t2}%
\end{table*}%

\begin{figure*}[!htbp]
	\centering
	\footnotesize
	\begin{minipage}{4cm}
		\centering
		\includegraphics[width=4.3cm]{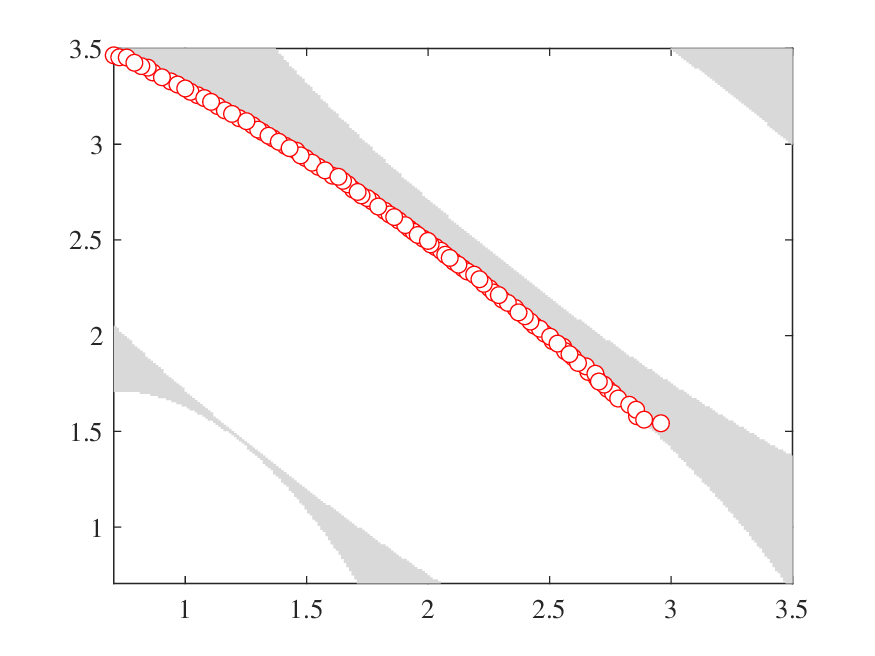}
		\text{(a)}
	\end{minipage}
	\begin{minipage}{4cm}
		\centering
		\includegraphics[width=4.3cm]{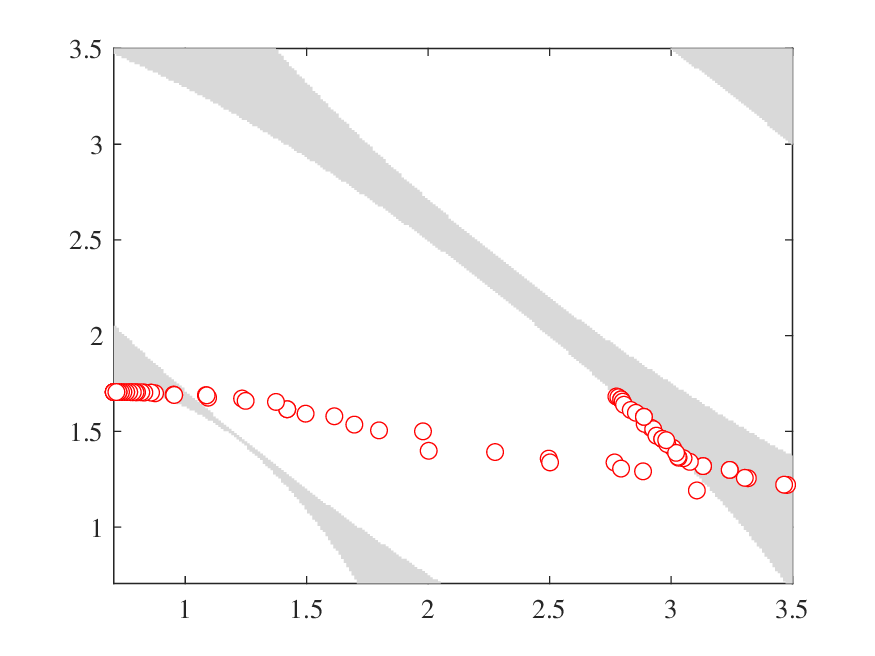}
		\text{(b)}
	\end{minipage}
	\begin{minipage}{4cm}
		\centering
		\includegraphics[width=4.3cm]{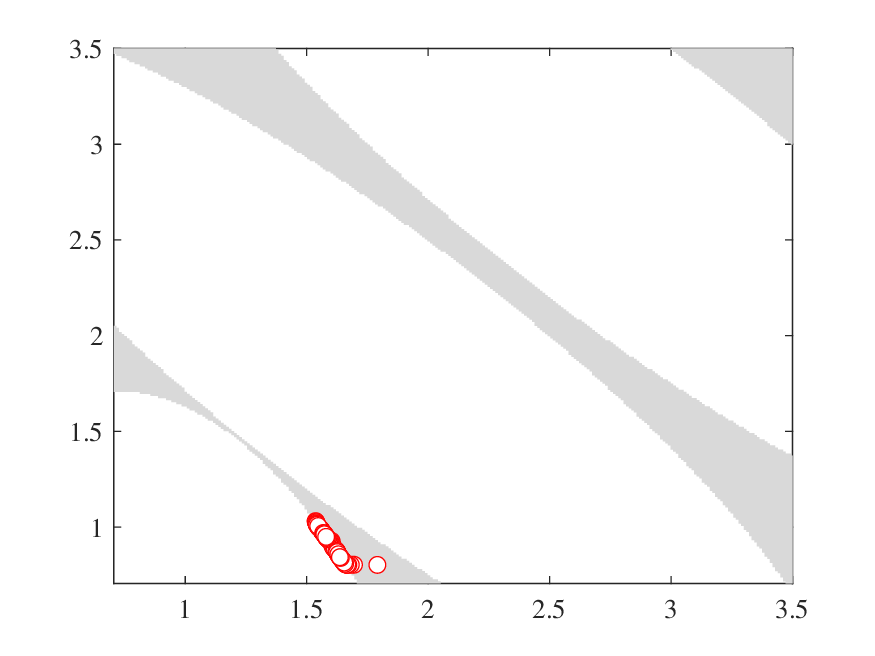}
		\text{(c)}
	\end{minipage}
	\begin{minipage}{4cm}
		\centering
		\includegraphics[width=4.3cm]{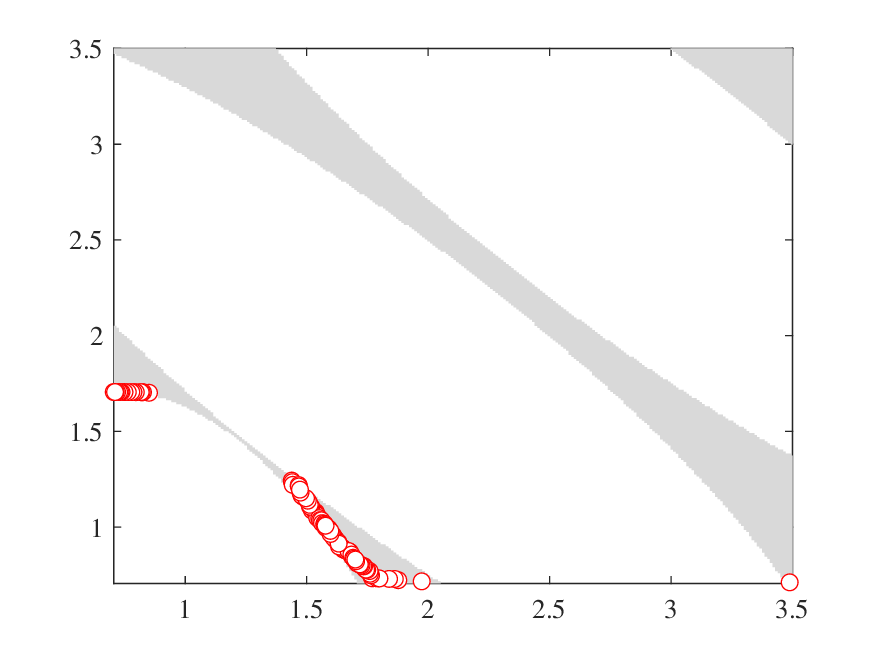}
		\text{(d)}
	\end{minipage}
	\begin{minipage}{4cm}
		\centering
		\includegraphics[width=4.3cm]{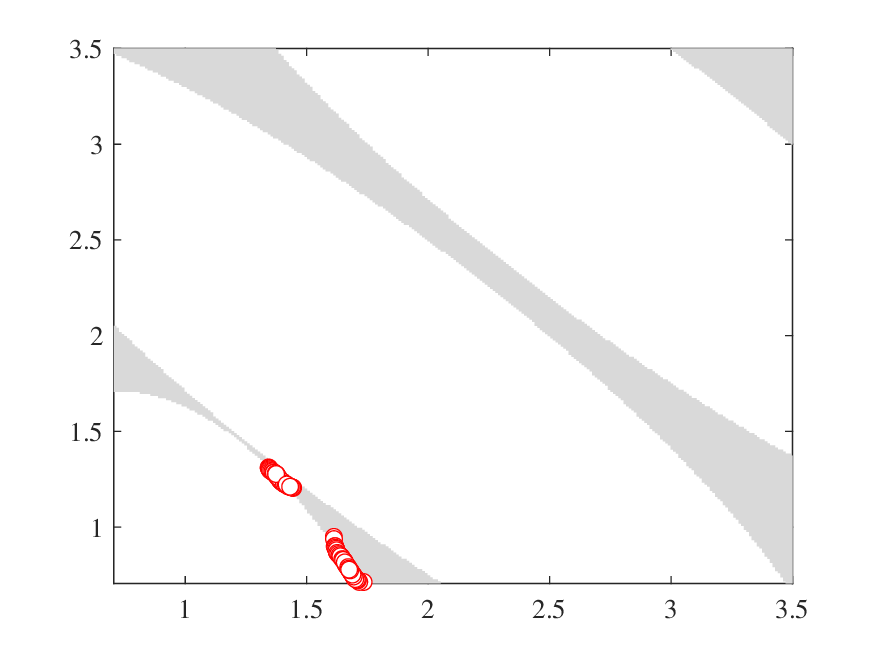}
		\text{(e)}
	\end{minipage}
	\begin{minipage}{4cm}
		\centering
		\includegraphics[width=4.3cm]{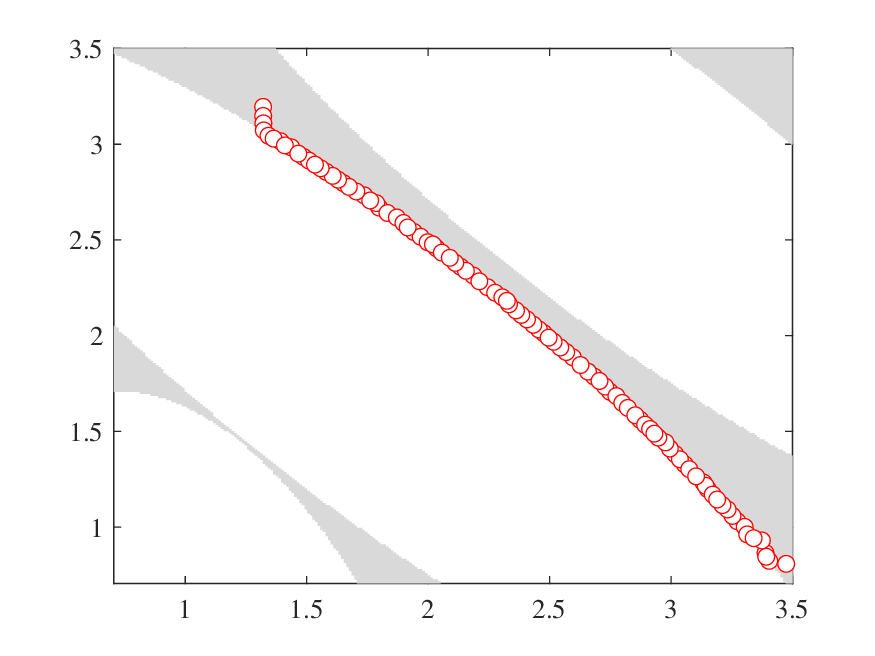}
		\text{(f)}
	\end{minipage}
	\begin{minipage}{4cm}
		\centering
		\includegraphics[width=4.3cm]{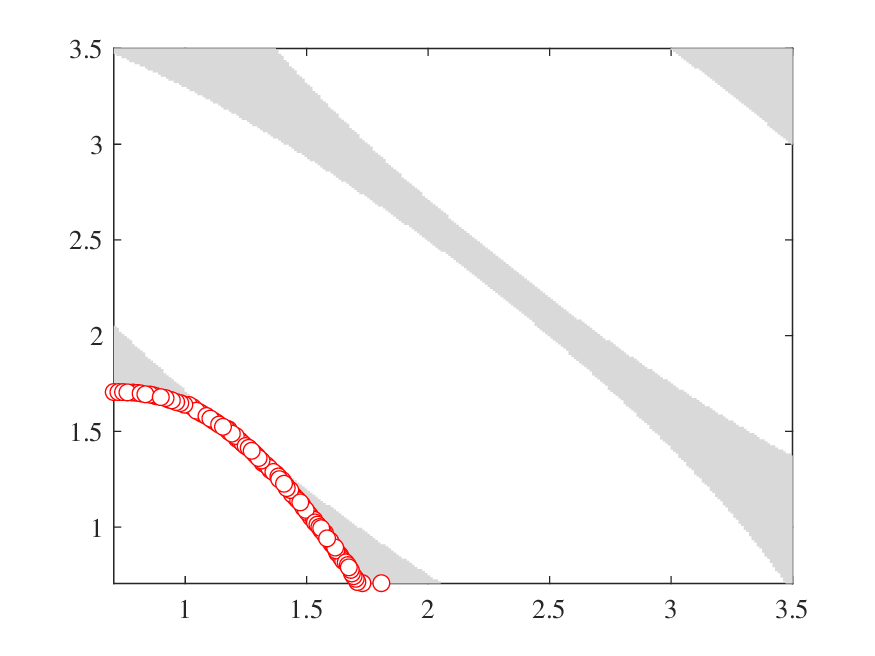}
		\text{(g)}
	\end{minipage}
	
	\caption{Solutions with the median IGD obtained by (a) CMOEA-MS, (b) MSCMO, (c) cDPEA, (d) MCCMO, (e) EMCMO, (f) IMTCMO, (g) DRMCMO on LIRCMOP/BC6. The gray area represents feasible region. }
	\label{f8}
\end{figure*}

LIRCMOP is mainly designed to test the ability of algorithms to cross large infeasible regions and also covers a variety of complex constraint scenarios. Based on LIRCMOP, the LIRCMOP/BC version binarizes the return values of all constraints; that is, the constraints will not give precise and quantifiable results but only provide binary results. This change increases the difficulty of algorithm processing. Table \ref{t2} lists the average IGD and standard deviation obtained by the proposed DRMCMO and other compared CMOEAs after 30 independent runs on the LIRCMOP/BC and DASCMOP/BC test suites. DRMCMO achieved the best IGD performance on 22 problems. Compared with CMOEA-MS, MSCMO, MCCMO, cPDEA, EMCMO, and IMTCMO, DRMCMO significantly outperformed the compared algorithms on 22, 19, 18, 20, 19 and 20 problems. Fig. \ref{f8} shows the results obtained by all algorithms on LIRCMOP/BC6. The results clearly show the significant gap between other algorithms and DRMCMO, among which only the final population of DRMCMO successfully covers the entire CPF. It is worth noting that both CMOEA-MS and IMTCMO are trapped in local optimality and fail to cross large infeasible regions effectively. This is mainly because when the constraints can only provide binary feedback, the effect of the fitness function that depends on the CV is significantly reduced, resulting in the performance of CMOEA-MS being limited. At the same time, $\varepsilon$-based methods relied on in IMTCMO is also completely ineffective, failing to properly relax the restrictions on the feasible region, which in turn causes the population to fall into local optimality. In the comparison of IGD convergence curves of all algorithms on LIRCMOP/BC5 in Fig. \ref{IGDC} (a), it can be seen that DRMCMO also has the fastest convergence speed. In order to further demonstrate the competitiveness and superior performance of DRMCMO, we performed a Friedman test on LIRCMOP/BC. In the average Friedman ranking results of all algorithms shown in Fig. \ref{f7}, DRMCMO performed best with a ranking of 1.46 and significantly outperformed other algorithms. 

\begin{figure*}[!htbp]
	\centering
	\footnotesize
	\begin{minipage}{4cm}
		\centering
		\includegraphics[width=4.3cm]{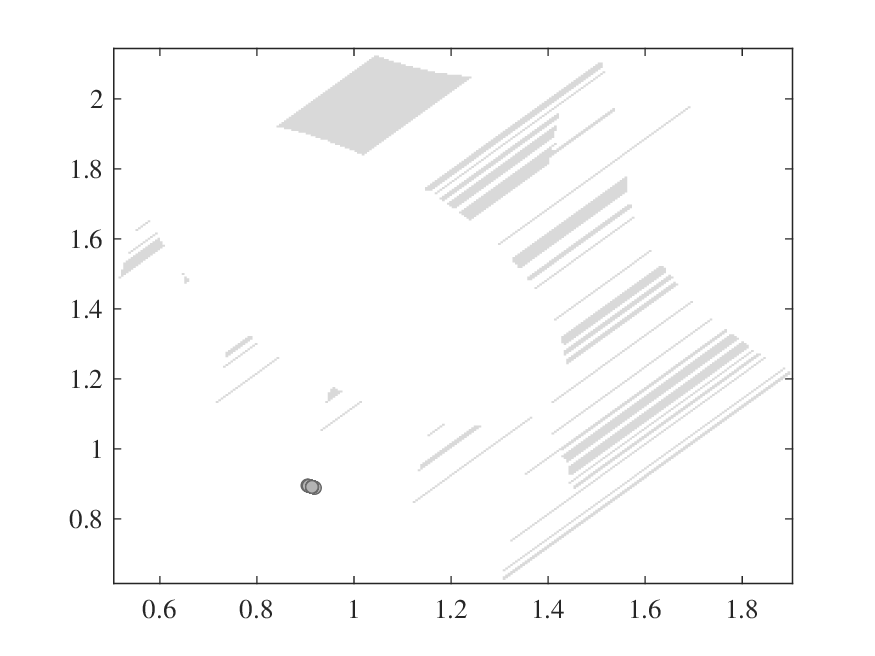}
		\text{(a)}
	\end{minipage}
	\begin{minipage}{4cm}
		\centering
		\includegraphics[width=4.3cm]{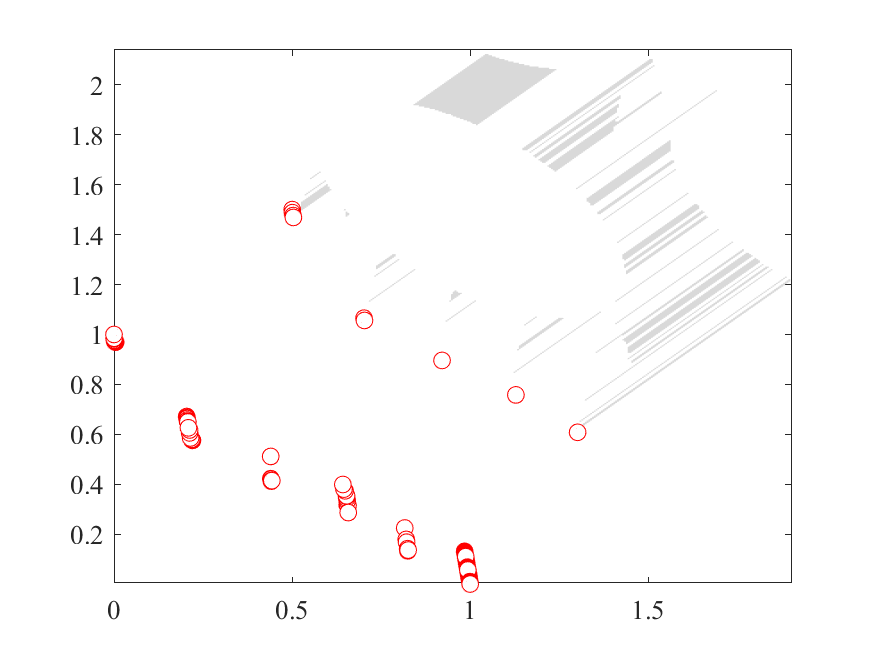}
		\text{(b)}
	\end{minipage}
	\begin{minipage}{4cm}
		\centering
		\includegraphics[width=4.3cm]{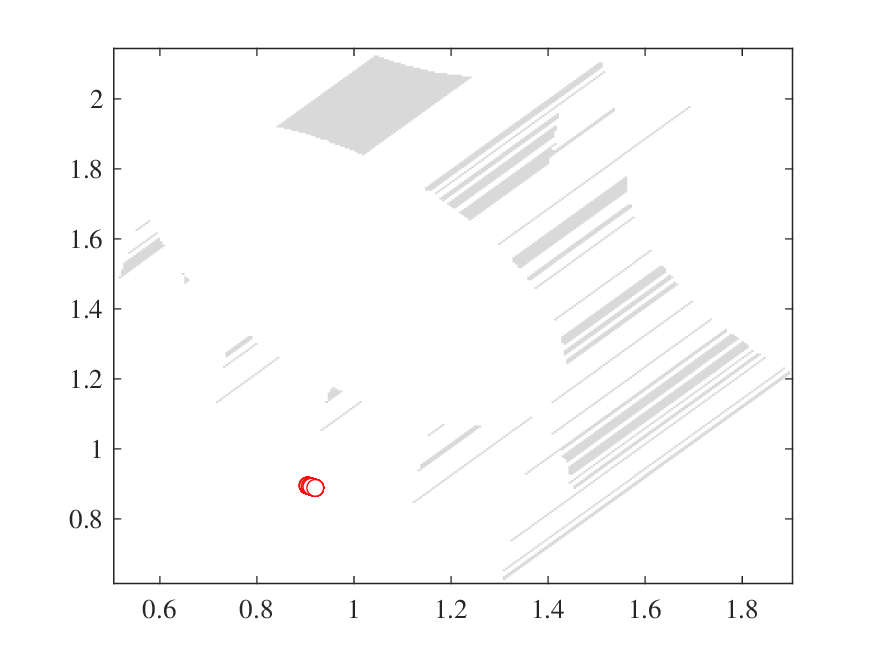}
		\text{(c)}
	\end{minipage}
	\begin{minipage}{4cm}
		\centering
		\includegraphics[width=4.3cm]{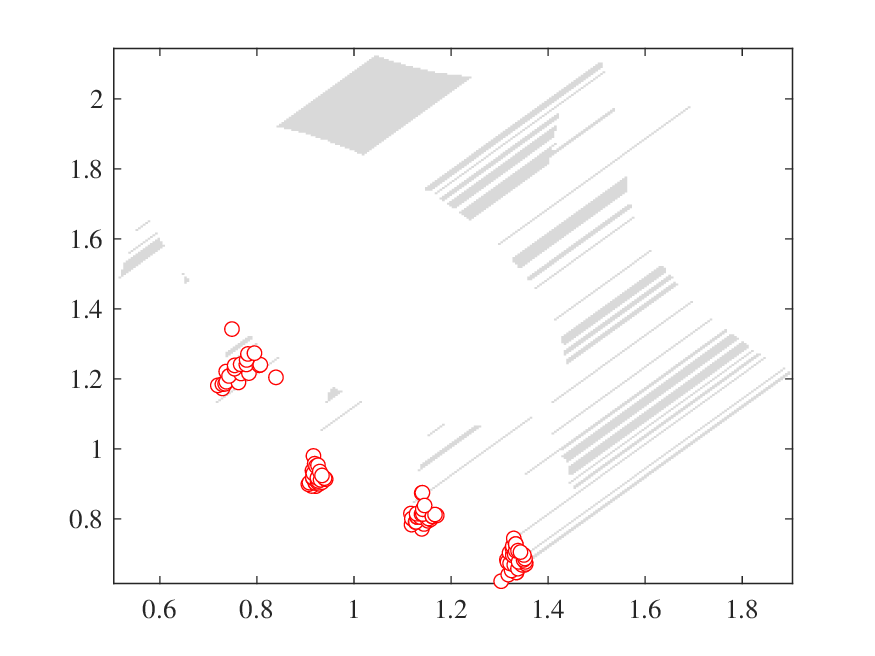}
		\text{(d)}
	\end{minipage}
	\begin{minipage}{4cm}
		\centering
		\includegraphics[width=4.3cm]{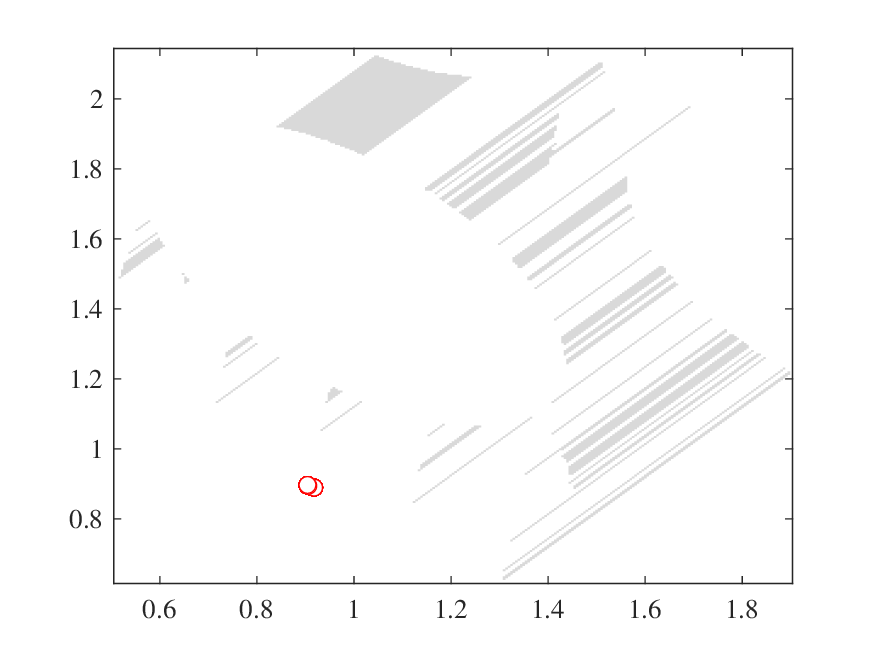}
		\text{(e)}
	\end{minipage}
	\begin{minipage}{4cm}
		\centering
		\includegraphics[width=4.3cm]{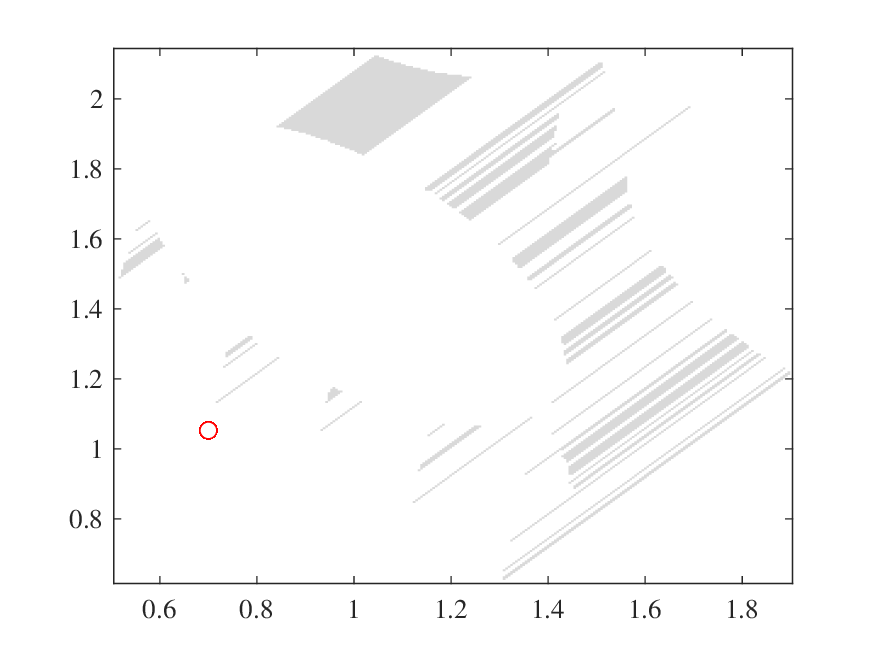}
		\text{(e)}
	\end{minipage}
	\begin{minipage}{4cm}
		\centering
		\includegraphics[width=4.3cm]{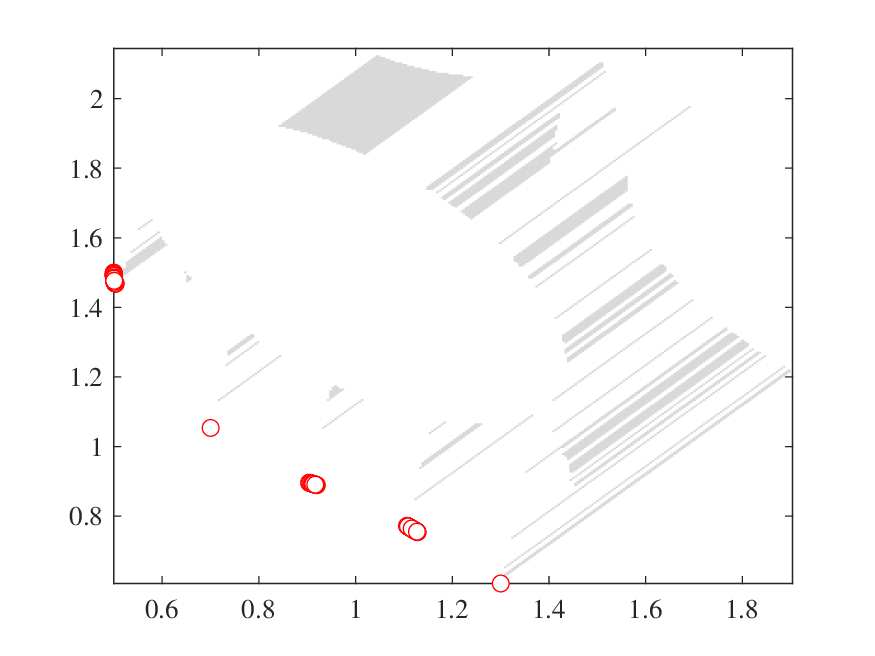}
		\text{(e)}
	\end{minipage}
	
	\caption{Solutions with the median IGD obtained by (a) CMOEA-MS, (b) MSCMO, (c) cDPEA, (d) MCCMO, (e) EMCMO, (f) IMTCMO, (g) DRMCMO on DASCMOP/BC3. The gray area represents feasible region. }
	\label{f10}
\end{figure*}

The DASCMOP test suite also covers a variety of complex constraint scenarios. Based on the original DASCMOP problem, we propose the DASCMOP problem with binary constraints (DASCMOP/BC). From the DASCMOP/BC experimental comparison data in Table \ref{t2}, we can see that DRMCMO performs significantly better than other algorithms. Fig. \ref{f10} shows the populations obtained by all algorithms on the DASCMOP/BC3 problem, among which only DRMCMO successfully searched the complete CPF. Due to the weakening of CV guidance and the failure of  $\varepsilon$-based methods, some algorithms such as CMOEA-MS, cDPEA, EMOCMO, and IMTCMO have fallen into local optima. Although MSCMO and MCCMO have roughly searched CPF, these two algorithms need to consume many evaluation times to maintain performance. Under the 100000 evaluation limit set in this paper, their performance is significantly inferior to the proposed DRMCMO. In the comparison of IGD convergence curves for all algorithms on DASCMOP/BC2, as shown in Fig. \ref{IGDC} (b), it is evident that DRMCMO exhibits the fastest convergence speed. Further analysis, judging from the average Friedman ranking results of all algorithms shown in Fig. \ref{f7}, DRMCMO performed the best with a ranking of 1.22 and was significantly better than other algorithms. Comprehensive experimental results on these three CMOP/BC test problem sets have proven that compared with existing top CMOEAs, our proposed DRMCMO has higher adaptability and competitiveness in processing CMOP/BC.

 \begin{figure}[!htbp]
	\centering
	\footnotesize
	\begin{minipage}{4.3cm}
		\centering
		\includegraphics[width=4.3cm]{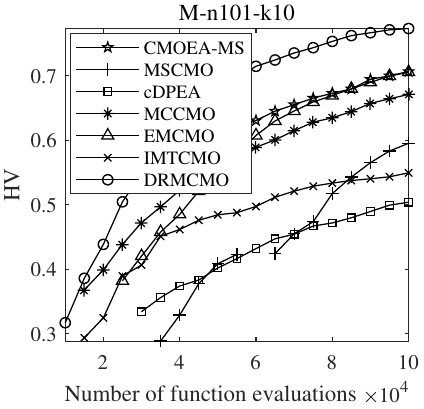}
		\text{(a)}
	\end{minipage}
	\begin{minipage}{4.3cm}
		\centering
		\includegraphics[width=4.3cm]{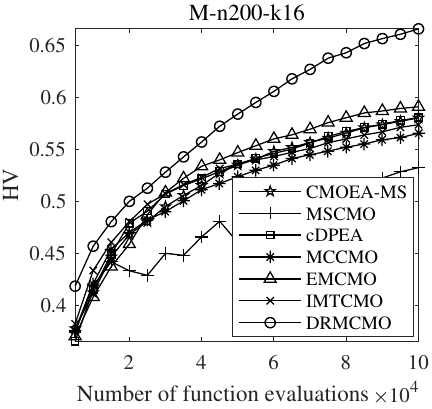}
		\text{(b)}
	\end{minipage}
	\centering
	\caption{HV curves from 30 independent runs of all algorithms on CVPR
		problems. (a) M-n101-k10 and (b) M-n200-k16 .}
	\label{HVC}
\end{figure}

\begin{figure}[!htbp]
	\centering
	\footnotesize
	\includegraphics[width=6cm]{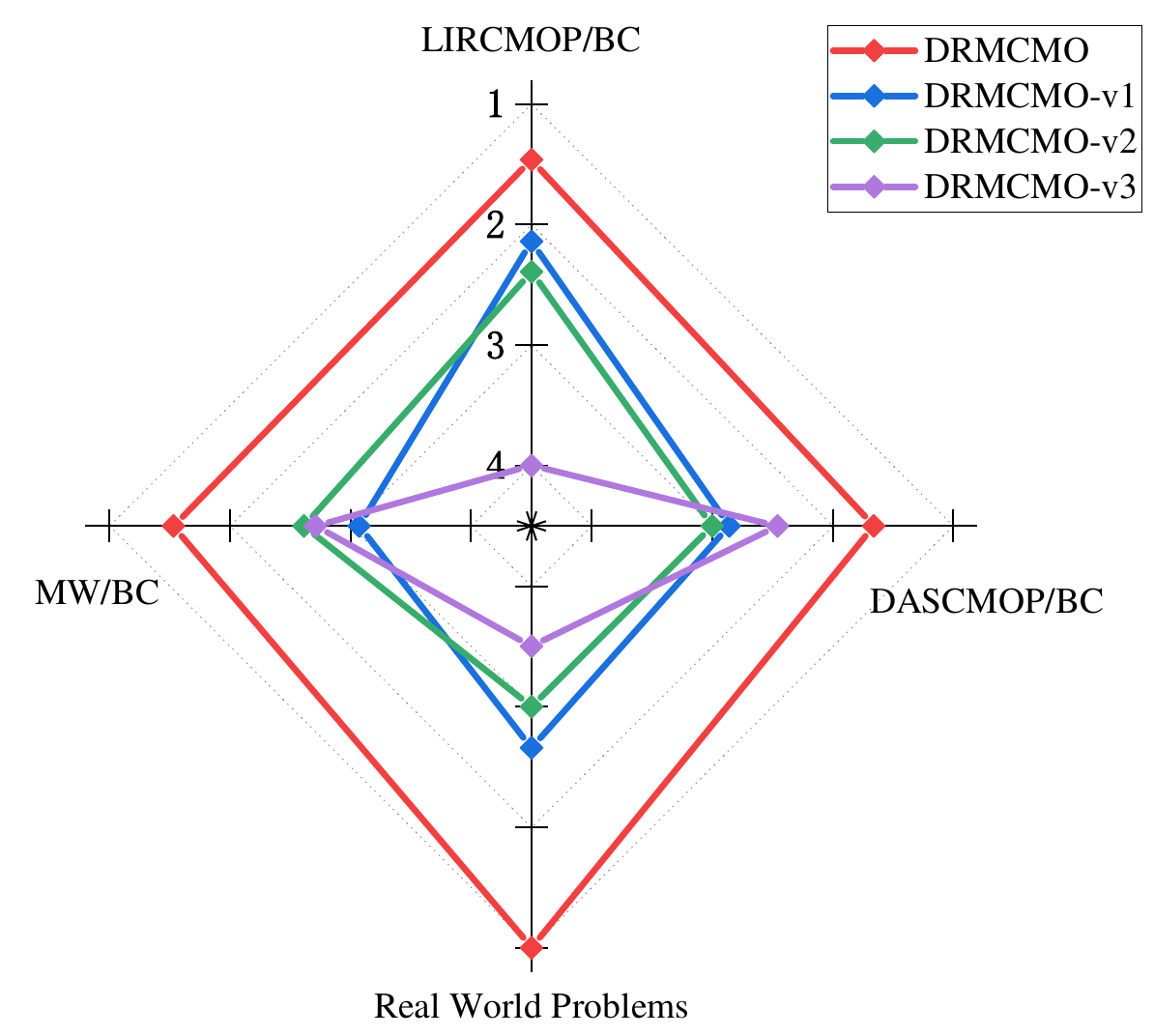}
	\centering
	
	\caption{ Friedman ranking among DRMCMO and its variants on MW/BC, LIRCMOP/BC, DASCMOP/BC, and real world problems.  }
	\label{f9}
\end{figure}

\begin{table*}[htbp]
	\centering
	\setlength{\tabcolsep}{3pt}
	\caption{HV  Results Obtained by the Proposed DRMCMO and All the Compared CMOEAs  on three Real-World problems. }
	\begin{threeparttable}
	{\fontsize{6.58pt}{9.5pt}\selectfont
	\begin{tabular}{cccccccr}
		\toprule
		\multicolumn{1}{c|}{\multirow{2}[4]{*}{Real world Problem (CVRP)}} & \multicolumn{2}{c|}{Multi-Stage CMOEAs} & \multicolumn{2}{c|}{Multi-Population CMOEAs} & \multicolumn{2}{c|}{Multi-Task CMOEAs} & \multicolumn{1}{c}{Proposed CMOEA} \\
		\cmidrule{2-8}    \multicolumn{1}{c|}{} & CMOEA\_MS & \multicolumn{1}{c|}{MSCMO} & cDPEA & \multicolumn{1}{c|}{MCCMO} & EMCMO & \multicolumn{1}{c|}{IMTCMO} & \multicolumn{1}{c}{DRMCMO} \\
		\midrule
		M-n101-k10 & 7.0640e-1 (2.68e-2) - & 5.9518e-1 (3.57e-2) - & 5.0391e-1 (3.74e-2) - & 6.7104e-1 (4.31e-2) - & 7.0558e-1 (1.13e-2) - & 5.4947e-1 (4.33e-2) - & \multicolumn{1}{c}{\hl{7.7305e-1 (1.28e-2)} }\\
		M-n121-k7 & 6.6955e-1 (2.45e-2) $\approx$ & 5.7814e-1 (9.92e-2) - & 5.4695e-1 (4.33e-2) - & 6.7884e-1 (3.28e-2) $\approx$ & 6.7427e-1 (3.23e-2) $\approx$ & 6.7243e-1 (2.62e-2) $\approx$ & \multicolumn{1}{c}{\hl{6.7982e-1 (1.48e-2)} }\\
		M-n151-k12 & 4.2189e-1 (2.81e-2) - & 3.8793e-1 (3.90e-2) - & 2.8303e-1 (4.93e-2) - & 3.7663e-1 (2.77e-2) - & 4.3674e-1 (2.48e-2) - & 3.8742e-1 (2.93e-2) - & \multicolumn{1}{c}{\hl{4.9489e-1 (1.11e-2)} }\\
		M-n200-k16 & 5.8252e-1 (1.12e-2) - & 5.4066e-1 (3.56e-2) - & 5.8719e-1 (1.06e-2) - & 5.8703e-1 (1.81e-2) - & 6.0836e-1 (1.52e-2) & 6.0259e-1 (1.71e-2) - & \multicolumn{1}{c}{\hl{6.0930e-1 (1.08e-2)} }\\
		M-n200-k17 & 5.7777e-1 (1.36e-2) - & 5.3970e-1 (1.70e-2) - & 5.7732e-1 (1.63e-2) - & 5.7322e-1 (1.31e-2) - & 5.9275e-1 (5.64e-3) - & 5.7941e-1 (1.52e-2) - & \multicolumn{1}{c}{\hl{6.0225e-1 (9.71e-3)} }\\
		\midrule
		+/-/$\approx$ & 0/4/1 & 0/5/0 & 0/5/0 & 0/4/1 & 0/3/2 & 0/4/1 &  \\
		\bottomrule
	\end{tabular}%
	}
	\begin{tablenotes}
		\item The best result for each row is highlighted.``N/A'' means that no feasible solution can be found. ``+'', ``-'' and ``$\approx$'' indicate that the result is significantly \\better, significantly worse, and  statistically similar to the results obtained by DRMCMO,respectively.
	\end{tablenotes}
	\end{threeparttable}
	\label{t3}%
\end{table*}%

\subsection{Ablation Study Results Analysis}

To further evaluate the effectiveness of the proposed DRMCMO, we designed and conducted a series of ablation experiments. In these experiments, three variants of DRMCMO were used to explore the role of key components in the algorithm: 
\begin{enumerate}[]
	\item  DRMCMO-v1:  removes the center point shifting method to assess its contribution to overall algorithm performance.
	\item DRMCMO-v2: The parameter \(\alpha\), which controls the variation of the detection region radius \(r\), is modified to use the following linear change method:
\begin{equation}
	\label{e14}
	\text{Linear}: \alpha = \frac{\textit{k} - \textit{k}_\textit{s}}{\textit{K} - \textit{k}_\textit{s}}
\end{equation}
The purpose of this modification is to evaluate the effectiveness of \(\alpha\) compared to the original S-shaped variation. The original method for changing \(\alpha\) is presented in equation \ref{e10}.

	\item DRMCMO-v3: completely replaces the DRM with the CDP to verify the overall effectiveness of DRM.
\end{enumerate}
By comparing the performance degradation of these variant algorithms, we gain deeper insights into the impact of different components of DRM on the algorithm's overall performance. DRMCMO and its variants were thoroughly tested across all benchmark problems.  The experimental results, illustrated in the average Friedman ranking in Fig. \ref{f9}, show that the original DRMCMO significantly outperforms the three variants, indicating that the original composition of the algorithm is optimal. The removal of any component negatively impacts performance. Notably, DRMCMO-v3 received the lowest performance ranking, underscoring the critical role of DRM in enhancing the algorithm's effectiveness.

\subsection{Experimental Results on real-world applications}
To further verify the effectiveness of our proposed algorithm, we applied it to real-world optimization problems for testing. The test scenario we chose is the capacitated vehicle routing problem (CVRP) \cite{altabeeb2019improved}, a logistics optimization problem. The goal is to determine the optimal route so that a group of vehicles with limited cargo capacity starts from a central point, serves multiple customers, and returns to the starting point while minimizing the total driving distance or cost as much as possible. In addition, some scholars also use multiobjective optimization to solve the CVRP problem \cite{jingjing2022cluster}, namely MO-CVRP. This paper adopts the problem model of Wang et al. \cite{wang2023co}. Formally, MO-CVRP is defined as follows:
\begin{equation}
Min\begin{cases}F_1=&\sum_{i=0}^{N_{c}}\sum_{j=0}^{N_{c}} x_{ij}c_{ij}\\F_2=&\sum_{v=1}^{M_{v}}(d_m-\bar d)^2/{M_{v}}\end{cases}
\end{equation}
where $F_1$ represents the total mileage of the vehicle, $N_{c}$ is the number of customers, $c_{ij}$ is the distance between the $i$th customer and the $j$th customer; when $i = 0$, $c_{0j}$ represents the distance from the warehouse to the $j$th customer, when $j = 0$, $c_{i0}$ represents the distance from the $i$th customer to the warehouse, and $x_{ij}$ represents whether there is a path between the $i$th customer and the $j$th customer, where $x_{ij}$ = 1 means there is a path, otherwise $x_{ij}$ = 0. $F_2$ is the load variance between vehicles, where $M_{v}$ is the number of vehicles used, $d_m$ represents the load of the m-th vehicle, and $\overline{d}$ is the average load of $M_{v}$ vehicles. In addition, the MO-CVRP needs to satisfy seven rule-based constraints. The specific equation definitions can be found in the original paper \cite{wang2023co}. In this paper, all rule-based constraints in MO-CVRP can only return satisfied or unsatisfied. The problem scenario dataset we use is the widely used CVRP dataset "Classic M" \cite{uchoa2017new} \cite{christofides1981exact}. This dataset contains 5 problems. Taking "M-n101-k10" as an example, "n101" represents 101 customers, and "k10" represents 10 vehicles. The number of evaluations for all problems is set to 10000 times, and the population size is set to 100. Table \ref{t3} shows the HV results of our DRMCMO and other comparison algorithms on this real problem. It can be seen that DRMCMO achieves the best results on all problems and significantly outperforms other algorithms. In the comparison of HV convergence curves shown in Fig. \ref{HVC}, DRMCMO exhibits the fastest convergence speed. Additionally, the Friedman ranking in Fig. \ref{f7} shows that DRMCMO surpasses other leading CMOEAs by a substantial margin.  The experimental results fully demonstrate the competitiveness of our proposed DRMCMO in handling real-world CMOP/BC.

\section{Conclusion and Future Work}
\label{s6}
Many optimization scenarios in the real world need to be modeled as CMOP/BC, especially optimization problems with many rule-based constraints (safety standards, map restrictions, operational restrictions, etc.). However, research on CMOP/BC is almost blank in the evolutionary multiobjective optimization research community. Existing CMOEAs face many challenges in dealing with such problems. Given this, this paper proposes a new algorithm, DRMCMO, based on the detection region method and specifically designed for the unique challenges of CMOP/BC problems. We conduct comprehensive experimental verification in multiple series of test problems and real-world problem scenarios. The experimental results show that DRMCMO significantly outperforms existing CMOEAs in performance, demonstrating its effectiveness and competitiveness in dealing with CMOP/BC. 

Nevertheless, DRMCMO is not impeccable. In particular, when facing CMOP/BC with CPF and PF separated and tiny feasible regions, due to the weakening of the guiding role of CV, DRMCMO mainly relies on CDP to find feasible solutions in the early stage and sometimes may not be able to search for feasible solutions. This problem highlights the limitations of DRMCMO and points out one of the challenges that future researchers in the field of CMOP/BC need to face. Future research can further explore strategies to improve the algorithm's performance under extreme constraints, thereby solving the CMOP/BC problem more comprehensively.

In addition, real-world optimization scenarios often encompass binary constraint problems and more complex challenges, such as large-scale decision variables, many objectives, and dynamic constraints. While the DRMCMO presented in this paper is not designed to address these issues directly, it aims to provide a novel solution to the CMOP/BC that has not been fully explored. Fortunately, the DRMCMO  exhibits high portability and can be integrated with other algorithms with distinct characteristics and advantages, allowing it to tackle a broader range of problems. This adaptability represents a promising direction for future research.

\section*{Acknowledgment}
This work was supported by the Scientific Research Project of Xiang Jiang Lab (22XJ02003), the University Fundamental Research Fund (23-ZZCX-JDZ-28), the National Science Fund for Outstanding Young Scholars (62122093), the National Natural Science Foundation of China (72421002), the National University of Defense Technology Youth Innovation Science Fund Project (ZK25-62), the science and technology innovation Program of Hunan Province (ZC23112101-10), and the Hunan Natural Science Foundation Regional Joint Project (2023JJ50490). 
The authors would like to thank the support by the COSTA: complex system optimization team of the College of System Engineering at NUDT.


%

%


\ifCLASSOPTIONcaptionsoff
  \newpage
\fi



\bibliography{reference}
\end{document}